\documentclass{article}

\pdfoutput=1
\PassOptionsToPackage{numbers, compress}{natbib}
\usepackage[final]{neurips_2023}

\usepackage[utf8]{inputenc} % allow utf-8 input
\usepackage[T1]{fontenc}    % use 8-bit T1 fonts
\usepackage{hyperref}       % hyperlinks
\usepackage{url}            % simple URL typesetting
\usepackage{booktabs}       % professional-quality tables
\usepackage{amsfonts}       % blackboard math symbols
\usepackage{nicefrac}       % compact symbols for 1/2, etc.
\usepackage{microtype}      % microtypography
\usepackage{xcolor}         % colors

\usepackage{amsmath,bm}
\usepackage{enumitem}
\usepackage{pifont}
\usepackage{multirow}
\usepackage{makecell}
\usepackage{graphicx}
\usepackage{colortbl}
\usepackage{color}
\usepackage{wrapfig}
\usepackage{graphicx}
\usepackage{tabularx}

\usepackage{capt-of}

\title{Open Visual Knowledge Extraction via Relation-Oriented Multimodality Model Prompting}

\author{%
Hejie Cui$^1$\footnotemark[1] \quad Xinyu Fang$^2$\thanks{These authors contributed equally to this work.} \quad Zihan Zhang$^2$ \quad Ran Xu$^1$ \quad Xuan Kan$^1$ \quad Xin Liu$^3$ \\ \textbf{Yue Yu}$^4$ \quad \textbf{Manling Li}$^5$ \quad \textbf{Yangqiu Song}$^3$ \quad \textbf{Carl Yang}$^1$$\thanks{Correspondence to: j.carlyang@emory.edu}$\\
$^1$Emory University \quad $^2$Tongji University \quad $^3$ The Hong Kong University of Science and Technology \\ $^4$ Georgia Institute of Technology \quad $^5$ Northwestern University \\
}

\definecolor{maroon}{cmyk}{0,0.1,0.01,0.01}
\definecolor{blue}{cmyk}{0.95,0.0,0.2,0.2}
\definecolor{yellow}{cmyk}{0.01,0.0,0.2,0.01}
\definecolor{lightblue}{cmyk}{0.1,0.0,0.02,0.02}
\definecolor{case_verb}{HTML}{fbde84}
\definecolor{case_adj}{HTML}{cccdff}
\definecolor{case_noun}{HTML}{bfeaf1}
\definecolor{case_ff}{HTML}{e65352}
\definecolor{case_error}{HTML}{ffff00}

\usepackage{xspace}
\usepackage[T1]{fontenc}  

\newcommand{\ours}{{\fontfamily{cmtt}\selectfont {OpenVik}}\xspace}
\newcommand{\cmmnt}[1]{}

\begin{document}
\maketitle

\begin{abstract}
  % Traditional knowledge extraction has predominantly centered on textual data. The integration of visual information into interconnected knowledge extraction can offer distinct insights that complement the textual modality, facilitating a more comprehensive understanding of concepts and mutually reinforcing intricate semantic structures with multiple information sources. 
  Images contain rich relational knowledge that can help machines understand the world. Existing methods on visual knowledge extraction often rely on the pre-defined format (e.g., sub-verb-obj tuples) or vocabulary (e.g., relation types), restricting the expressiveness of the extracted knowledge.
  In this work, we take a first exploration to a new paradigm of open visual knowledge extraction.
  To achieve this, we present \ours which consists of an open relational region detector to detect regions potentially containing relational knowledge and a visual knowledge generator that generates format-free knowledge by prompting the large multimodality model with the detected region of interest. We also explore two data enhancement techniques for diversifying the generated format-free visual knowledge.
  % Extensive evaluations, including generation-metric-based performance, multifaceted human assessments, and the comparison against both established explicit knowledge graphs and implicit knowledge in large language models highlight the correctness and uniqueness of the extracted open visual knowledge. Moreover, the integration of this extracted knowledge across various visual reasoning tasks shows consistent improvements, indicating the real-world applicability of our proposed method. 
  Extensive knowledge quality evaluations highlight the correctness and uniqueness of the extracted open visual knowledge by \ours. Moreover, integrating our extracted knowledge across various visual reasoning applications shows consistent improvements, indicating the real-world applicability of \ours.
\end{abstract}

\section{Introduction}
\label{sec:intro}
% 1. VKE is important, yet existing work does pre-defined format and vocabulary
Knowledge extraction has been widely studied on texts~\cite{cui2022can,alani2003automatic,fan2012automatic,cui2023survey} for enhancing logical reasoning~\cite{teru2020inductive,galkin2022inductive,chen2022fuzzy} and explainable AI~\cite{holzinger2018current,yu2021sumgnn,chan2021salkg,xu2022counterfactual}, and recent studies have explored \textit{open} knowledge extraction through categorizing seed relations~\cite{zhang2020aser,sap2019atomic} and eliciting from language models~\cite{wang2020language}.
% Knowledge extraction has been widely studied~\cite{} and recent studies have explored open extraction through ... ~\cite{} and eliciting from language models~\cite{shin-etal-2020-autoprompt}.
Visual knowledge extraction, on the other hand, captures intricate details like tools, sizes, and positional relationships, which are often difficult to express exhaustively in texts~\cite{sadeghi2015viske,li2020gaia,wang2021visual,cui2023pv2tea}. 
Yet existing approaches of visual knowledge extraction are either restricted by a fixed knowledge format~\cite{xu2017scene,zellers2018neural,densecap,kim2019dense} or the predefined sets of objects/relations~\cite{xu2017scene,zellers2018neural,kan2021zero}.
% Existing work has explored different knowledge formats to extract visual knowledge from images, e.g., scene graph~\cite{xu2017scene,zellers2018neural,he2022towards}, object-centric~\cite{densecap} and SVO-based caption~\cite{kim2019dense}.
% For example, scene graph generation~\cite{xu2017scene,zellers2018neural} locate objects in the image and identify visual predicates between subjects and objects in the form of (s, v, o) triples; image captioning~\cite{vinyals2015show,xu2015show,you2016image} holistically describe the content of an image with a single sentence; dense captioning~\cite{densecap} predict a set of object-centric descriptions across regions; relational dense captioning~\cite{kim2019dense}; a recent work OpenSGG~\cite{he2022towards} also extends traditional scene graph generation to open-vocabulary objects, where the model is designed to predict visual relations for unseen objects. 
% Yet they are either restricted by a fixed knowledge format or the predefined sets of objects/relations.
While efficient at capturing interactions between objects, the produced visual knowledge is often limited in richness and confined to a single format, falling short in representing the diverse real-world information that can be complemented by visual data.

% Recently, large language and multimodality models have exhibited remarkable successes in capturing commonsense knowledge across various tasks, especially facilitating few-shot~\cite{gao2021making,min2022noisy,lang2022co} and zero-shot learning~\cite{meng2022generating,kojima2022large,zhou-etal-2022-prompt}. 
% The potential of prompt-based learning for pre-trained vision-language models~\cite{alayrac2022flamingo,radford2021learning,Su2020VL-BERT} has been explored for handling diverse data types across multiple modalities, such as images and texts, with improved performance in tasks including
% image classification~\cite{menon2023visual,zhou2022conditional}, segmentation \cite{luddecke2022image} and visual question answering~\cite{guo2022images}. 
% Leveraging the substantial information encapsulated within these pre-trained multimodality models to extract explicit visual knowledge can enrich existing resources, potentially laying the groundwork for advances in interpretability research and mitigating the hallucination issue often associated with large models~\cite{ji2023survey,dai-etal-2023-plausible}.
% More related to us, several works that adopt prompt-based learning to enable the model to generate responses based on the given images or videos~\cite{yang2022prompt,wang2022language,yang2023mm,hu2022promptcap}.

% 2. we explore format-free VKE. explain and compare with a pre-defined format
In this endeavor, we propose to further explore a new paradigm of open visual knowledge extraction (\ours). %we take an initial step to explore a new visual knowledge extraction paradigm: open visual knowledge extraction.
% Traditional methods often employ the sub-verb-obj format, which, while efficient at capturing interactions and relationships between objects, fails to encapsulate the richness of information that visual data holds.
Specifically, we propose to generate relation-oriented, but format-free knowledge that includes a wider variety of elements, such as descriptions, insertions, and attributes, among others. 
Drawing inspiration from the wealth of knowledge encapsulated in large models~\cite{DBLP:conf/nips/Wei0SBIXCLZ22,zelikman2022star,tsimpoukelli2021multimodal}, we propose to leverage pre-trained large multimodality models by eliciting open visual knowledge through relation-oriented visual prompting. 
This approach allows for a more nuanced understanding of visual data, mirroring how humans naturally emphasize certain aspects of visual scenes when perceiving and describing visual information, leading to more flexible visual knowledge extraction.
% or they grapple with the challenges posed by highly unbalanced long-tail training data distributions~\cite{tang2020unbiased}

% 3. we present \ours, explain two modules briefly
Our proposed \ours framework consists of two modules, an open relational region detector and a format-free visual knowledge generator.
% \manling{
It is a unique challenge to detect the regions potentially containing relational knowledge, since traditional region detectors primarily focus on learning predefined object classes. %The traditional object detector is not applicable since there are no predefined object classes for relational region detection. 
To learn the regression of relational regions, we propose to use free-form knowledge descriptions as supervision and leverage knowledge generation as a training objective. % To this end, we leverage knowledge generation as the training objective, where the free-form knowledge descriptions are used as supervision for the fine-tuned region of interest regression. 
With the detected regions, the remaining question is how to interpret these regions into free-form knowledge. We propose a visual knowledge generator by harnessing the power of language variety enhancement in large pre-trained multimodality models. Specifically, we prompt them to generate knowledge descriptions of any formats and condition the generation on the detected relational regions.  
% Drawing inspiration from the wealth of knowledge encapsulated in large pre-trained multimodality models \manling{repeated in the above paragraph}, we harness their power to fine-tune our visual knowledge extractor. 
% This extractor, conditioned on the detected relational regions, generates format-free knowledge that is complemented by language variety enhancement.
% }
% The visual knowledge generator enables open vocabulary knowledge discovery, we adopt an open-ended encoder-decoder knowledge generator that generates a flexible number of knowledge descriptions based on soft visual feature prompting; 

% 4. we develop two data enhancement techniques
\cmmnt{describe the challenge first, pitch it as a unique challenge for the new paradigm} 
% To overcome the limitation of biased training data and promote diverse, meaningful knowledge discovery, we design two data enhancement strategies to incorporate external knowledge graphs and large language models in assisting relation recognition and boosting entity perception as well as promoting within-sequence and between-sequence language variation. 
However, establishing a new paradigm of open visual knowledge extraction is challenging due to the absence of comprehensive and diverse training data. Existing datasets sources such as scene graphs~\cite{xu2017scenegraph, krishna2017visual}, dense captions~\cite{densecap}, and dense relational subsets~\cite{kim2019dense} often exhibit a long-tail distribution biased to more prevalent relations and entities~\cite{tang2020unbiased}. Brute-force merging of these datasets could exacerbate the distribution bias inherent in the data. To alleviate the bias, we propose two diversity-driven data enhancement strategies based on an adapted TF-IDF+ score, involving random dropping and data augmentation with external knowledge resources. These strategies optimize data distributions and richness, thus fostering diverse open visual knowledge extraction.

% 5. exp
% contribution: 1. format-free paradigm with a new model 2. data enhancement techniques 3. exp
% In summary, the main contributions of this work are three-fold:
% \begin{itemize}[nosep,leftmargin=*]
% \item We propose a new open paradigm and approach for visual knowledge extraction, \ours, composed of an open relational region detector and a format-free visual knowledge generator by prompting a large multimodality model. 
% %The proposed approach transcends the limitations of fixed knowledge formats and vocabularies, better aligning with the narrative convention used by humans. 
% % \item To amplify knowledge diversity and mitigate the constraints of training data bias, we devise two data enhancement techniques, which enhance both relation recognition and entity perception.
% \item We devise data enhancement techniques that amplify the knowledge diversity of \ours.
% \item Comprehensive evaluations are conducted to gauge the quality and utility of the extracted visual knowledge, encompassing knowledge generation performance, multi-faceted knowledge assessment, and the comparison with existing knowledge sources. Besides, applications on visual reasoning downstream tasks including text-to-image retrieval, grounded situation recognition, and visual commonsense reasoning demonstrate the practicality of \ours.
% \end{itemize}
We implement extensive evaluations to assess the quality and utility of the open visual knowledge extracted by \ours, encompassing: 1) directly evaluating the performance of knowledge generation; 2) engaging human evaluators for a multi-faceted assessment of in-depth knowledge quality; and 3) comparing the open visual knowledge extracted with \ours with existing knowledge sources, such as non-parametric knowledge from the ConceptNet knowledge graph, and parametric knowledge from the GPT-3.5 large language model. 
Furthermore, the utility of the extracted open visual knowledge is validated through its integration with several common applications that require visual understanding, including text-to-image retrieval, grounded situation recognition, and visual commonsense reasoning.
% \xin{why to choose these downstream applications? I did not find the reason or connection in the introduction.}
These applications demonstrate consistent improvements, affirming the practical utility of \ours.

\vspace{-0.1cm}
\section{Related Work}
\label{sec:related}
\vskip -0.1em

\textbf{Visual knowledge extraction.}
% \cmmnt{just one paragraph for vke, focus on paradigm shift}
% Vision is a key modality for humans to understand the world. 
Recent advancements in knowledge extraction have extended from being purely text-driven to incorporating images~\cite{ding2022mukea,liu2019mmkg}. VisKE~\cite{sadeghi2015viske} is designed to verify relations between pairs of
entities, e.g., \textit{eat}(\textit{horse}, \textit{hay}). Scene graphs, which locate objects in the image and identify visual predicates between subjects and objects in a triple format, e.g., (\textit{man, on, chair}), are extensively studied for vision understanding~\cite{xu2017scene,zellers2018neural,zareian2020bridging}. A recent work OpenSGG~\cite{he2022towards} extends SGG to open-vocabulary objects, enabling the relation prediction for unseen objects. Other studies have explored caption-like formats, like dense captioning~\cite{densecap} with a set of object-centric descriptions across regions, and relational captioning~\cite{kim2019dense} focusing on relational information between objects. Despite these advancements, existing methods either adhere to a pre-defined format and vocabulary or are constrained by the biased distribution of training sets. This highlights the pressing need for a format-free approach in visual knowledge extraction with knowledge diversity.  
% has explored different knowledge formats to extract visual knowledge from images, e.g., scene graph~\cite{xu2017scene,zellers2018neural,he2022towards}, object-centric~\cite{densecap} and SVO-based caption~\cite{kim2019dense}.
% For example, scene graph generation~\cite{xu2017scene,zellers2018neural} locate objects in the image and identify visual predicates between subjects and objects in the form of (s, v, o) triples; image captioning~\cite{vinyals2015show,xu2015show,you2016image} holistically describe the content of an image with a single sentence; dense captioning~\cite{densecap} predict a set of object-centric descriptions across regions; relational dense captioning~\cite{kim2019dense}; a recent work OpenSGG~\cite{he2022towards} also extends traditional scene graph generation to open-vocabulary objects, where the model is designed to predict visual relations for unseen objects. 
% \hejie{to do.}

\textbf{Large model prompting.}
% Large language models have demonstrated a surprising ability to 
% Prompt-based learning is a powerful paradigm in natural language processing that involves fine-tuning large language models using text prompts to achieve specific goals. This approach has shown remarkable success in various tasks for few-shot~\cite{gao2021making,min2022noisy,lang2022co} and zero-shot learning~\cite{meng2022generating,kojima2022large,zhou-etal-2022-prompt}. 
% Multimodal prompts have been introduced to exploit the potential of prompt-based learning for pretrained vision-language models~\cite{alayrac2022flamingo,radford2021learning,Su2020VL-BERT} in handling diverse data types across multiple modalities, such as images and text, with improved performance in a wide range of tasks including
% image classification~\cite{menon2023visual,zhou2022conditional}, segmentation \cite{luddecke2022image} and visual question answering~\cite{guo2022images}.
% More related to us, several works that adopt prompt-based learning to enable the model to generate responses based on the given images or videos~\cite{yang2022prompt,wang2022language,yang2023mm,hu2022promptcap}. Despite their great potential, these language models have been observed to generate unfaithful content when given prompts, a phenomenon known as "hallucination"~\cite{ji2023survey,dai-etal-2023-plausible}. Depending solely on these models may result in an imprecise knowledge graph which may also hurt downstream performances. \yue{Optional: add a few sentences for your method.}
Recently, large language and multimodality models have exhibited remarkable successes in capturing commonsense knowledge across various tasks, especially facilitating few-shot~\cite{gao2021making,xu2023neighborhood,lang2022co,yu2022cold} and zero-shot learning~\cite{kojima2022large,zhou-etal-2022-prompt,yu2023large}. 
The potential of prompt-based learning for pre-trained vision-language models~\cite{alayrac2022flamingo,radford2021learning,Su2020VL-BERT} has been explored for handling diverse data types across multiple modalities, such as images and texts, with improved performance in tasks including
image classification~\cite{menon2023visual,zhou2022conditional}, segmentation \cite{luddecke2022image} and visual question answering~\cite{guo2022images}. 
Leveraging the substantial information encapsulated within these pre-trained multimodality models to extract explicit knowledge can enrich existing resources, potentially laying the groundwork for advances in interpretability research and mitigating the hallucination issue associated with large models~\cite{ji2023survey,dai-etal-2023-plausible}.
% More related to us, several works that adopt prompt-based learning to enable the model to generate responses based on the given images or videos~\cite{yang2022prompt,wang2022language,yang2023mm,hu2022promptcap}.
% Prompting large language models:
%  prompt a multimodal caption model
% prompting large language models -> prompt a multimodal caption model

\section{Method}
\vspace{-0.1cm}
\label{sec:method}
In this section, we introduce our new paradigm and two key model design novelty featuring \ours, relation-oriented multimodality model prompting and diversity-driven data enhancement. 
\begin{figure}[ht]
    \vskip -1.0em
    \centering
    \includegraphics[width=1.0\linewidth]{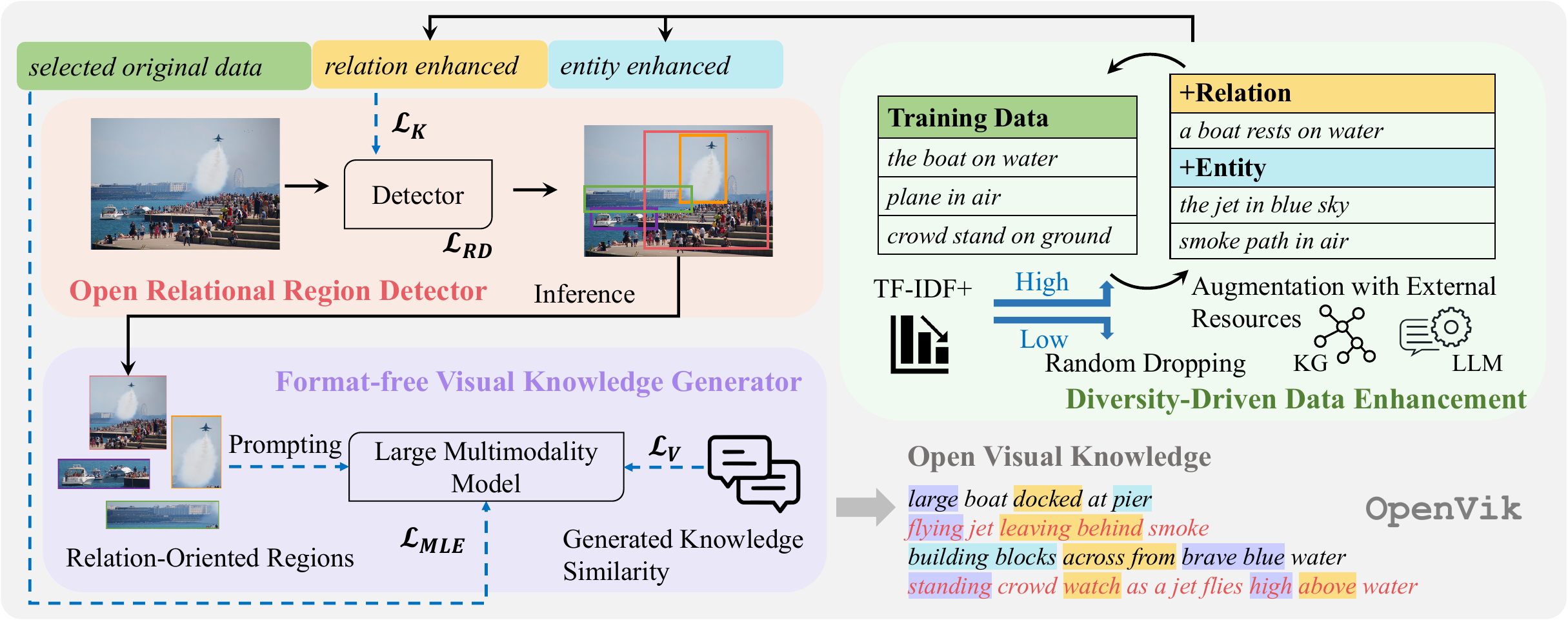}
    \caption{The overview of \ours. The left orange and purple panels illustrate key components of relation-oriented multimodality model prompting: open relational region detector and format-free visual knowledge generator. The right green one depicts diversity-driven data enhancement strategy. \ours is designed to extract relation-oriented {\color{case_ff}format-free} open visual knowledge with novel {\colorbox{case_noun}{entities}}, diverse {\colorbox{case_verb}{relations}}, and nuanced {\colorbox{case_adj}{descriptive details}}.}
    % \manling{the arrow on the top should point to both ``relation enhanced'' and ``entity enhanced''}
    \vskip -1.5em
    \label{fig:overall}
\end{figure}

\subsection{Open Visual Knowledge Extraction}
Given a dataset $\mathcal{D}=\left\{\left(\mathcal{I}_i, \mathbf{T}_i, \mathbf{U}_i\right)\right\}_{i=1}^M$ consisting of $M$ samples, $\mathcal{I}_i$ is the $i$-th image (such as the input image in Figure \ref{fig:overall}), $\mathbf{T}_i=\left\{\mathcal{T}_j\right\}_{j=1}^{n_i}$ is a set of $n_i$ region descriptions (such as ``\textit{the boat on water}'' in Figure \ref{fig:overall}), $\mathbf{U}_i=\left\{\mathcal{U}_j\right\}_{j=1}^{n_i}$ is the set of $n_i$ relation-oriented visual regions, where each $\mathcal{T}_j$ corresponds to a visual region $\mathcal{U}_j \in \mathbf{U}_i$ in image $\mathcal{I}_i$. 
% \manling{I feel here some example input/output will be appreciated. We can use Figure 1. Normally I will add one example in the Introduction (like Figure 4) to explain the input/output and what is free-form knowledge, and then use it as the walk-through example in the method section. Since there is no space, adding an example output description into Figure 1 will be also appreciated, and use Figure 1 as the walk-through example in the method section.}
The goal of our open visual knowledge discovery is to train a model $\mathcal{M}$ capable of producing a set of format-free knowledge descriptions (such as ``\textit{large boat docked at pier}'' in Figure \ref{fig:overall}) given any image $\mathcal{I}_k$ during the inference stage.

% \paragraph{Relational Region Detector}
% \paragraph{Open Visual Knowledge Generator}

% \subsection{\ours: Architecture and Training}
\subsection{Relation-Oriented Multimodality Model Prompting}
% \subsection{Relation-Oriented Conditional Generation}
% \subsection{Training and Inference Process.}

The overall architecture of \ours is shown in Figure \ref{fig:overall}. It comprises two modules: an open relational region detector $\mathcal{M}_v$ and a format-free visual knowledge generator $\mathcal{M}_t$.
The two modules are learned separately during training with our diversity-enhanced data (Section \ref{sec:data}) and combined to produce format-free visual knowledge at inference. Specifically, the relational region detector $\mathcal{M}_v$ takes an image $\mathcal{I}_i$ as the input and learns to select a flexible number of relational regions $\mathbf{U}_i=\left\{\left(\mathcal{U}_j\right)\right\}_{j=1}^{n_i}$ that captures object interactions, each corresponding to a description $\mathcal{T}_j$ in $\mathbf{T}_i$; the visual knowledge generator $\mathcal{M}_t$ generates format-free knowledge descriptions by prompting and fine-tuning the multimodality model with the guidance of detected visual region $\mathcal{U}_j$. All notations for the region detector and knowledge generator are detailed in Table \ref{tab:hype4detector} and Table \ref{tab:hype4generator}, respectively.

\begin{minipage}{\textwidth}
\begin{minipage}[t]{0.47\textwidth}
\makeatletter\def\@captype{table}
\captionof{table}{Notations for open region detector.}
\label{tab:hype4detector}
\resizebox{\linewidth}{!}{%
\begin{tabular}{c|c}
\toprule
\textbf{Notation} & \textbf{Meaning} \\
\midrule
$\mathcal{I}_i$ & input image of the relational region detector\\
$\mathcal{U}_j$ & relation-centric box label \\
$\mathbf{U}_i$ & set of relation-centric boxes of an image \\ 
$\mathcal{T}_j$ & region description of a box\\
$\mathbf{T}_i$ & set of region descriptions of the an image \\ 
$\mathcal{L}_\text{RD}$ & region regression loss supervised by union regional boxes\\
$\mathcal{L}_\text{K}$ & knowledge generation loss supervised by GT relational knowledge  \\
$\mathcal{L}_v$ & the overall objective of the relational region detector \\
\bottomrule 
\end{tabular}
}
\end{minipage}
\qquad
\begin{minipage}[t]{0.47\textwidth}
\makeatletter\def\@captype{table}
\captionof{table}{Notations for knowledge generator.}
\label{tab:hype4generator}
\resizebox{\linewidth}{!}{%
\begin{tabular}{c|c}
\toprule
\textbf{Notation} & \textbf{Meaning} \\
\midrule
$\mathcal{I}_i$ & the input image of the knowledge generator\\
$T_a, T_b$ & two regional knowledge descriptions of one same image \\
$N_i$ & the number of generated knowledge descriptions of an image \\
% $s (T_a, T_b)$ & the semantic cosine similarity of two input generated descriptions \\
$\phi$ & hyper-parameter controlling the penalty slightly different sequences \\
$\mathcal{L}_\text{MLE}$ & the language modeling loss of the generation decoder \\
$\mathcal{L}_{\text{V}}$ & inter-sequence information variety regularizer \\
$\alpha$ & weight hyper-parameter balancing generation accuracy and variety \\
$\mathcal{L}_l$ & the overall objective of the knowledge generator \\
\bottomrule 
\end{tabular}
}
\end{minipage}
\end{minipage}

\vspace{0.2cm}

\textbf{Open relational region detector.}
Although existing object detection algorithms have been widely recognized for their efficiency in object detection, they are usually restricted to object-centric visual regions in a predefined set, and thus cannot directly capture open relational information with a single box. Detecting regions containing relational knowledge remains to be a challenge. 
We make two adaptions on the object detection FasterRCNN~\citep{ren2015faster} to train the open relational region detector:
\begin{itemize}[nosep,leftmargin=*]
\item \textit{Region Regression}: we change the original object-centric region labels to our newly created relation-centric box labels, denoted as $\mathbf{U}_j$. The foreground of each relation-centric region label $\mathcal{U}_j$ is created by taking the union of the object-level bounding boxes of the entities, i.e., \textit{boat, water}, contained in a ground truth region knowledge description $\mathcal{T}_j$. This forms the region regression loss $\mathcal{L}_\text{RD}$.
\item \textit{Knowledge Supervision}: To assist with the refinement of the bounding box, we replaced the object-centric label classification in traditional object detectors with knowledge supervision. A pre-trained generator is finetuned to create the regional description grounded to the given region. This is supervised by the cross-entropy loss $\mathcal{L}_\text{K}$ with region description $\mathbf{T}_j$.
\end{itemize}

% Then we replace the label classification with description generation, where region descriptions $\mathbf{T}_i$ serve as supervision. 
% In addition, to further assist bounding box refinement, we add the ground truth relational knowledge as additional supervision $\mathcal{L}_\text{K}$, where a pre-trained generator \manling{what is the model architecture used in the detector? why there is an additional generator? How is the detector and the generator connected? Are the detector parameters learned from scratch? }is finetuned to generate the ground truth regional description $\mathcal{T}_j$ of the given region. 
The training objective $\mathcal{L}_l$ of the relational region detector is formulated as below, where $\mathcal{L}_\text{RD}$ is the regional regression loss and $\mathcal{L}_\text{K}$ is the knowledge supervision loss, 
\begin{equation}
\mathcal{L}_v = \mathcal{L}_\text{RD} + \mathcal{L}_\text{K}.
\end{equation}

\textbf{Format-free visual knowledge generator.}
OpenVik provides better knowledge grounding by conditioning the generator on the detected relational region, leading to a reasoning-driven generation.
Specifically, the detected bounding box (such as the box containing ``\textit{boat}'' and ``\textit{pier}'' on the far left) is utilized as a visual prompt when fine-tuning the visual knowledge generator. 
The model architecture of the knowledge generator is built upon a combined large multimodality model, which composes a pre-trained vision transformer ViT-B~\cite{DBLP:conf/iclr/DosovitskiyB0WZ21} and the image-grounded text decoder of BLIP~\citep{DBLP:conf/icml/0001LXH22}. The two modules are jointly trained on a generic image-text paired dataset comprising over 14 million entries and fine-tuned on the image captioning task, which delivered state-of-the-art performance. 

In our visual knowledge generator, the decoder takes the ViT visual representation of the entire image as input and leverages the detected regional mask as a binary visual prompt. This prompt aids in filtering out the background and directing attention toward the relational foreground. The generation of format-free knowledge from the decoder is supervised by the language modeling loss $\mathcal{L}_\text{MLE}$, which further refines visual attention during the knowledge generation process. As a result, our approach facilitates the production of format-free outcomes that extend beyond the conventional sub-verb-obj form.
% Each block is composed of a causal self-attention layer, a cross-attention layer, and a feed-forward network. 
%A \verb|[Decode]| token is used to signal the beginning of a sequence and an end-of-sequence token is used to signal its end. 
Besides, to improve information variety, we introduce an amplifying penalty factor for highly similar knowledge generation. For any two generated sequences $T_a$ and $T_b$ describing image $\mathcal{I}_i$,
\vspace{-0.1em}
\begin{equation}
\label{equ:diverse}
\mathcal{L}_{\text{V}}= \frac{1}{N_i} \sum_{N_i} \operatorname{ReLU}\left(-\log \left(1-\left(s\left(T_a, T_b\right)-\phi\right)\right)\right),
\end{equation}
\vskip -0.5em
where $N_i$ is the number of generated knowledge of image $\mathcal{I}_i$, $s\left(T_a, T_b\right)$ indicates the semantic cosine similarity, and $\phi$ is a hyper-parameter set as 0.01 controlling the penalty on sequences with only slight difference (e.g. ``\textit{dog chasing the man}'' and ``\textit{dog licking the man}'') to be relatively small. 

The training objective $\mathcal{L}_l$ of the format-free visual knowledge generator is formulated as 
\vspace{-0.1em}
\begin{equation}
\mathcal{L}_l = \alpha \times \mathcal{L}_\text{MLE} + \left ( 1-\alpha \right ) \times \mathcal{L}_\text{V},
\end{equation}
\vskip -0.5em
where $\alpha$ is a weighting hyper-parameter we set as 0.7. 
% \paragraph{Inference of Diverse Open Visual Knowledge Extraction.}
The trained relational region detector and visual knowledge generator are combined during inference. Given any image $\mathcal{I}$, the open relational region detector first detects a flexible number of open relations regions of interest, then each detected region $\mathcal{R}$ is passed to the format-free visual knowledge generator, where a relation-oriented format-free knowledge phrase (such as ``\textit{flying jet leaving behind smoke}'' in Figure \ref{fig:overall}) is generated to describe the given visual focus subarea $\mathcal{R}$ of the image. To further encourage within-sequence language variety during inference, we leverage the contrastive decoding strategy from~\cite{su2022contrastive}, which improves over nucleus sampling and beam search.

\subsection{Diversity-driven Data Enhancement}
\label{sec:data}
The training data for relational knowledge extraction usually exhibits a long-tail distribution, where more prevalent but simple relations such as \textit{in}, \textit{on}, and \textit{wear} dominate the training set~\cite{tang2020unbiased}. Consequently, the model trained with such a biased dataset may render limited, and repetitive knowledge. As a remedy, we propose two data enhancement techniques to optimize the data distribution. 
As the foundational measure for given relation $r$'s importance, we design a grid TF-IDF+ score $\mathcal{S}_{r}$ ~\cite{paik2013novel,xu2023weakly}:
\vspace{-0.2em}
\begin{equation}
\mathcal{S}_{r} = (\log(\frac{N}{1+\mathit{f}_r * \alpha_{1}}))^{\alpha_{2}},
\end{equation}
\vskip -0.5em
where $N$ is the total number of knowledge phrases in the datasets, $\mathit{f}_r$ is the number of occurrences of the relation $r$, $\alpha_{1}$ and $\alpha_{2}$ are the grid scales whose values are selected based on $\mathit{f}_r$. 

\textbf{Random dropping on low-quality data.}
We first remove repeated knowledge descriptions in the same image and then randomly drop descriptions that contain frequently occurring yet meaningless relations with a low $\mathcal{S}_{r}$ (e.g., ``\textit{people on ground}'') from the original dataset. % while augmenting the rare but meaningful ones
Specifically, if the $\mathcal{S}_{r}$ of the relation in a description is relatively low, i.e., 0.4, we remove it at a random dropping rate of 0.5. This process repeats for all descriptions in an image until the remaining set is 0.6 times the size of the original training set. Consequently, the training data bias is mitigated by removing low-quality data. 
% \\
% \xinyu{probably a vague expression will be better(and list the hyperparameter in the appendix): Specifically, if the $\mathcal{S}_{r}$ of the relation in a description is relatively low, we remove it at a random dropping rate. This process repeats for all descriptions in an image many times until the diversity and fullness of the remaining set can be balanced. }

\textbf{Data augmentation with external knowledge resources.}
For the relations with high TF-IDF+ scores, we leverage external knowledge resources from both non-parametric (i.e., ConceptNet~\cite{speer2017conceptnet}) and parametric (i.e., COMET~\cite{DBLP:conf/acl/BosselutRSMCC19}) knowledge resources to promote diverse knowledge generation~\cite{yang2020generative}.
\noindent\underline{\textit{\ding{52} Enhance Relation Recognition}}: 
For each training description with a high TF-IDF+ score, we perform semantic parsing to get all the objects and complement additional relations (e.g., ``\textit{rest}'' in Figure \ref{fig:overall}) between each pair of them by mapping the nodes and retrieving edges from the ConceptNet. Each retrieved knowledge triplet is converted to a knowledge phrase and added to the training set for generator training.
With this introduced external knowledge, the knowledge generator ultimately yields a more robust and detailed representation of the underlying visual information of objects. This, in turn, bolsters the relation recognition of the visual knowledge generator.
\noindent\underline{\textit{\ding{52} Boost Entity Perception}}: 
For the description with the highest-scored TF-IDF+ relation given each image, we also leverage ConceptNet to enrich similar objects (e.g., ``\textit{jet}'') to the original object (e.g., ``\textit{plane}''). Additionally, we further introduce new entities (e.g., ``\textit{smoke}'' in Figure \ref{fig:overall}) and attribute descriptions (e.g., ``\textit{blue}'') by prompting the pre-trained attribute commonsense branch of the COMET model (Refer to Appendix \ref{app:data_aug} for more details). The entity-based enrichment potentially helps in boosting entity understanding and at the same time enhances the occurrence of important but rare relations in the training set.

% \subsection{Experimental Setup}
\subsection{Implementation Details}
\label{sec:exp-setup}

% \paragraph{Dataset}
% dataset statistics
Our training data are built based on Visual Genome~\cite{krishna2017visual} and its relation-enhanced version Dense Relational Captioning~\cite{kim2019dense}. 
% The dataset \ran{which dataset are you referring to? The Dense Relational Captioning dataset?} includes various relational information for each image that has been extracted from the original Visual Genome dataset. 
Each sample includes an image identified by a unique ID and a set of relational descriptors describing interactions among objects in the image. Specifically, each relational descriptor includes the full description text, the subject and object names contained in the description text, the relation between them, as well as the bounding box coordinates of the subject and object.
% The dataset information is summarized in Table \ref{tab:dataset}, where \textbf{split} specifies the dataset split, \textbf{\#image} indicates the number of images included in the dataset, \textbf{\#triplet} shows the number of relational information or "triplets" extracted from the images, \textbf{\#predicate} displays the total number of unique predicates obtained after deduplication, and \textbf{\#node} implies the total number of subjects and objects for the specific dataset type.
The dataset statistic information is summarized in Table \ref{tab:dataset} in the Appendix \ref{app:data}. 
%It provides a rich source of information for exploring the relationships between objects in images and serves as a valuable resource for various computer vision tasks such as image captioning, visual question answering, and scene understanding.

% \paragraph{Implementation Details}
% Our model is implemented in PyTorch~\citep{paszke2019pytorch} and trained on two Quadro RTX 8000 GPUs. The open relational region detector is initialized from the ResNet50-FPN backbone, then finetuned for another 20 epochs using the relational region bounding box with an AdamW optimizer~\citep{loshchilov2019decoupled} 
% and a weight decay of 0.9. The batch size is set as 4. The learning rate is warmed-up to 1e-3 and decayed linearly with a rate of 0.85. 
% The hidden dimension of the visual representation is set as 4096, and 512 for the bounding box. 
% The model detects a maximum of 30 bounding boxes for each image with the highest confidence to avoid misleading noises.
% The format-free visual knowledge generator is initialized from $\text{BLIP}_\text{base}$ with the basic ViT-B/16. The generator model is finetuned for 20 epochs with the same optimizer as the region detector. 
% % The learning rate was warmed-up to 1e-5 and decayed linearly with a rate of 0.85. 
% During inference, we take random image crops of resolution 384x384. The beam width is limited to 3 and the max sequence length to 20. 
% % model size needs to be specified

Our model is implemented in PyTorch~\citep{paszke2019pytorch} and trained on two Quadro RTX 8000 GPUs. The open relational region detector is initialized from the ResNet50-FPN backbone, then finetuned for another 20 epochs with the relational bounding box. The model detects a maximum of 30 bounding boxes for each image with the highest confidence to avoid misleading noises.
The format-free visual knowledge generator is initialized from $\text{BLIP}_\text{base}$ with the basic ViT-B/16 and finetuned for 20 epochs. Full details on learning parameters can be referred to in Appendix~\ref{app:imple}.
\section{Evaluation}
\label{sec:evaluation}
% In this section, we aim to assess the \textit{quality}, \textit{utility}, and \textit{effectiveness} of the proposed format-free visual knowledge extraction method by addressing four key research questions:
% \begin{itemize}[nosep,leftmargin=*]
% \item \textbf{RQ1}: How does the proposed \ours perform as a knowledge generation method, and how does the quality of our format-free visual knowledge compare with that generated by existing methods? 
% \item \textbf{RQ2}: How does the extracted format-free visual knowledge from \ours differentiate from the non-parametric knowledge found in Knowledge Graphs (KGs) and the parametric knowledge derived from Large Language Models (LLMs)?
% \item \textbf{RQ3}: Can the format-free visual knowledge extracted through \ours be utilized or integrated to improve downstream reasoning tasks? 
% \item \textbf{RQ4}: How effectively do the diversity regularizer $\mathcal{L}_{\text{V}}$ and the diversity-driven data enhancement strategies employed in \ours contribute to improving the quality of extracted knowledge?  
% \end{itemize}

In this section, we directly evaluate the extracted open visual knowledge from \ours from two perspectives: (1) knowledge generation performance with traditional generative metrics and in-depth knowledge quality assessment; (2) comparison with existing knowledge sources. Besides, ablation studies are conducted to study the influence of diversity design on the generated knowledge and data.  
% (1) How does the proposed \ours perform as a knowledge generation method, and how does the quality of our format-free visual knowledge compare with that generated by existing methods. (2) How does the extracted format-free visual knowledge from \ours differentiate from the non-parametric knowledge found in Knowledge Graphs (KGs) and the parametric knowledge derived from Large Language Models (LLMs)? 
% (3) How effectively do the diversity regularizer $\mathcal{L}_{\text{V}}$ and the diversity-driven data enhancement strategies employed in \ours contribute to improving the quality of extracted knowledge?  

%%%%%%%%%%%%%%%%%%%%%%%%%%%%%%%%%%%%%%%%%%%%%%%%%%%%%%%%%%%%%%%%%%%%%%%%%%%%
% \subsection{RQ1: Knowledge Generation Quality Evaluation}
\subsection{Evaluation on Generated Knowledge}

\begin{table*}[h]
\centering
% \small
\vskip -1.1em
\caption{Knowledge comparison of \ours and baselines on performance and in-depth quality (\%). }
\label{tab:performance}
\resizebox{1.0\linewidth}{!}{
\begin{tabular}{l|ccc|cccc}
\toprule
\multirow{2.5}{*}{\bf Method} &\multicolumn{3}{c|}{{\cellcolor{maroon}}{\bf Generation Performance}} &  \multicolumn{4}{c}{\cellcolor{lightblue}{\bf In-Depth Knowledge Quality}}\\
\cmidrule(lr){2-4} \cmidrule(lr){5-8} 
& {BLEU$\uparrow$} & {ROUGE-L$\uparrow$} & {METEOR$\uparrow$} &  {Validity$\uparrow$} & {Conformity$\uparrow$} & {Freshness$\uparrow$} & {Diversity$\uparrow$} \\
% \hline 
\midrule
% \multirow{3}{*}{\thead{Training Data}}
% \rowcolor{gray!15}\multicolumn{8}{l}{\emph{Training Data}}  \\
% \hline 
% \midrule
% Visual Genome~\cite{krishna2017visual} & --- & --- & --- &  --- & --- & --- & 0.675  \\
% Relational Caps~\cite{kim2019dense} & --- & --- & --- &  --- & --- & --- & 0.684 \\
% Diversity Enhanced & --- & --- &  --- &  --- & --- & --- &  \\
% \hline 
% \midrule
% \multirow{4}{*}{\thead{Scene Graph \\ Generation}}
\rowcolor{gray!15}\multicolumn{8}{l}{\emph{Closed/Open Scene Graph Generation}}  \\
% \hline 
IMP~\cite{xu2017scene} & 0.075 & 0.123 & 0.118 &  0.800 & 0.823 & 0.676 & 0.316 \\
Neural Motifs~\cite{zellers2018neural} & 0.229 & \underline{0.283} & \bf 0.273  & 0.822 & 0.767 & 0.667 & 0.349 \\
UnbiasSGG~\cite{tang2020unbiased} & 0.217 & 0.258 & 0.194 & 0.739 & 0.733 & 0.666 & 0.357\\
Ov-SGG~\cite{he2022towards} & 0.167 & 0.210 & 0.183 & 0.712 & 0.633 & 0.693 & 0.413 \\
% \hline 
% \midrule
% & GB-NET &  &  &  &  &  &  &  &  \\
% \multirow{2}{*}{\thead{Open Vocabulary SGG}}
% & Ov-SGG &  &  &  &  &  &  &  &  \\
% & CaCao &  &  &  &  &  &  &  &  \\
% \midrule
% \multirow{1}{*}{Dense Captioning} 
% \rowcolor{gray!15}\multicolumn{8}{l}{\emph{Dense Image Captioning}}  \\
% % \hline 
% DenseCap~\cite{densecap} & \underline{0.248} & 0.245 & 0.196  & 0.883 & 0.843 & \underline{0.790} & \underline{0.543} \\
% \hline
\midrule
% \multirow{1}{*}{Relational Captioning} 
\rowcolor{gray!15}\multicolumn{8}{l}{\emph{Dense Relational Captioning}}  \\
% \hline 
% & TSNet & 0.61 & 32.36 & 16.09 &  &  &  &  &  \\
% & MTTSNet & 0.88  & 34.27  & 18.73  &  &  &  &  &  &  \\
% & \multirow{2}{*}{MTTSNet+REM} & \multirow{2}{*}{1.12}  & \multirow{2}{*}{45.96}  & \multirow{2}{*}{0.184}  & \multirow{2}{*}{} & \multirow{2}{*}{} &  & \multirow{2}{*}{} & \multirow{2}{*}{} & \multirow{2}{*}{} \\ \\
MTTSNet+REM~\cite{kim2019dense} & 0.240 & 0.226 & 0.228  & \underline{0.897} & 0.852 &  0.754 & 0.375 \\
\midrule
\rowcolor{gray!15}\multicolumn{8}{l}{\emph{Region Captioning}}  \\
% \hline 
DenseCap~\cite{densecap} & 0.248 & 0.245 & 0.196  & 0.883 & 0.843 & 0.790 & 0.543 \\
Sub-GC~\cite{zhong2020comprehensive} & 0.272 & 0.263 & 0.221  & 0.892 & \underline{0.871} & \underline{0.795} & \underline{0.547} \\
BLIP~\cite{DBLP:conf/icml/0001LXH22} & 0.264 & 0.266 & 0.252  & 0.886 & 0.855 & 0.760 & 0.531 \\
BLIP2~\cite{li2023blip} & \underline{0.275} & \textbf{0.285} & \underline{0.257}  & 0.892 & \underline{0.871} & 0.766 & 0.535 \\
% \hline 
\midrule
% \multirow{1}{*}{Ours} 
\rowcolor{gray!15}\multicolumn{8}{l}{\emph{Open Visual Knowledge Extraction}}  \\
% \hline 
% \midrule 
\ours & \bf 0.280 & \underline{0.283} & 0.250 &  \bf 0.907 & \bf 0.883 & \bf 0.809 & \bf 0.619 \\
% 0.643
\bottomrule
\end{tabular}
}
\vskip -0.5em
\end{table*}

\textbf{\colorbox{maroon}{Generation performance.}}
To directly evaluate the visual knowledge generator, we compare the knowledge generated by \ours with a variety of baselines, including scene graph generation~\cite{xu2017scene, zellers2018neural, tang2020unbiased, he2022towards} (of which Ov-SGG employs an open vocabulary), dense relational captioning~\cite{kim2019dense}, and region captioning~\cite{densecap,zhong2020comprehensive,DBLP:conf/icml/0001LXH22,li2023blip}. Evaluation metrics are traditional language generation measures such as BLEU, ROUGE-L, and METEOR. The results, displayed in the left side of Table~\ref{tab:performance}, reveal that \ours outperforms captioning-based approaches and yields results on par with the best scene graph generation baseline. These findings underscore the effectiveness of the format-free visual knowledge generator through relation-oriented prompting of the large multimodality model.
% The baselines include:
% \begin{itemize}[nosep,leftmargin=*]
%     \item IMP: ~\cite{xu2017scene}
%     \item Neural Motifs: ~\cite{zellers2018neural}
%     \item UnbiasSGG: ~\cite{tang2020unbiased}
%     % \item GB-NET: ~\cite{zareian2020bridging}
%     \item Ov-SGG: ~\cite{he2022towards}
%     % \item TSNet: 
%     % \item MTTSNet: 
%     \item DenseCap: ~\cite{densecap}
%     \item MTTSNet+REM: ~\cite{kim2019dense}
% \end{itemize}

% design an additional metric based on the loss design of L_d (possibly SentenceBERT similarity among the training data, baselines, and our generation), also the traditional generation-based metrics for different baselines

% metrics
% MAP, METEOR, ..., 
% [Diversity Index]: definition and formula

% % baselines
% dense relational caption, 
% DRC's baseline (collect the results), 

% GT: [Diversity Index]
% Ours

%%%%%%%%%%%%%%%%%%%%%%%%%%%%%%%%%%%%%%%%%%%%%%%%%%%%%%%%%%%%%%%%%%%%%%%%%%%%
\textbf{\colorbox{lightblue}{In-depth knowledge quality.}}
% The design of evaluation metrics should correspond to the problems claimed in the introduction
To more thoroughly evaluate the quality and richness of the format-free visual knowledge extraction, beyond simply evaluating it as a language generation model with the limitation of training data, we incorporate four additional metrics~\cite{lu2022quark}, which delve into an in-depth quality evaluation of the extracted visual knowledge from four distinct perspectives:
\begin{itemize}[nosep,leftmargin=*]
    \item \textit{Validity} ($\uparrow$): whether the generated visual knowledge is valid to human.
    \item \textit{Conformity} ($\uparrow$): whether the generated knowledge faithfully depicts the scenarios in the images.
    \item \textit{Freshness} ($\uparrow$): the novelty of the knowledge, i.e., the proportion not present in the training set.
    \item \textit{Diversity} ($\uparrow$): the language variance between a randomly sampled pair of knowledge pieces.
    % \item \textbf{Evenness}: based on Pielou's evenness index, which is popular in environmental science to represent how equal the phrases in the overall knowledge triplets are.
\end{itemize}
% cite, we compute the score following this paper xxxx
% 3 persons: cohen's kappa, agreement
Among the four metrics, both the validity and conformity metrics involve human annotators. We randomly selected 100 images as the evaluative subset. Details regarding the scoring guidance and the interface provided to the annotators can be found in Appendix~\ref{app:human_eval}. 
%\hejie{The inter-annotator agreement was computed using Cohen's kappa, based on the ratings from three different evaluators.} 
The remaining metrics, i.e., freshness and diversity, are calculated automatically. 
The in-depth knowledge quality evaluation results are displayed in the right part of Table~\ref{tab:performance}, where the average pairwise Cohen's $\kappa$ on human evaluation results is 0.76 (good agreement).
The findings demonstrate that trained with the diversity-enhanced datasets, the format-free visual knowledge extracted by \ours significantly outperforms other types of baselines in terms of all four metrics.
The improvement of diversity, in particular, reaches 14\% relatively compared with the inference results from the second runner DenseCap, indicating the advantage of \ours in generating rich and comprehensive visual knowledge.

% \subsection{Knowledge Comparison Evaluation}
\subsection{Comparison with Existing Knowledge Sources}
\begin{figure}[h]
    \vskip -0.5em
    \centering
    \includegraphics[width=1.0\linewidth]{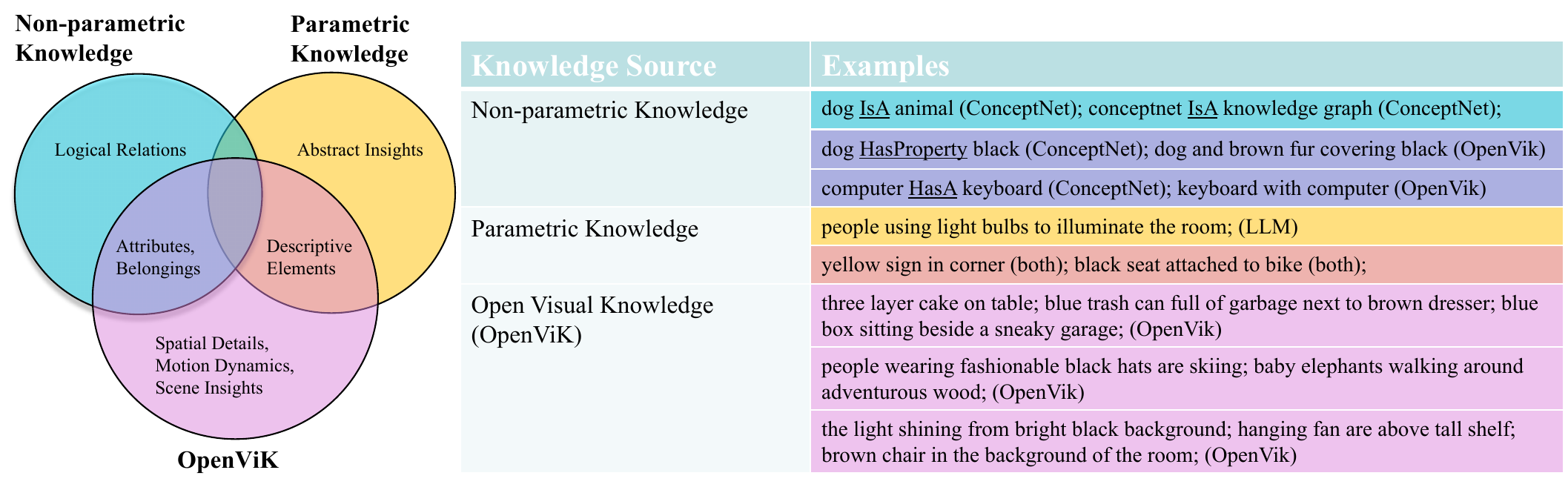}
    \vskip -0.8em
    \caption{The Venn diagram of knowledge comparison between the open visual knowledge from \ours with the non-parametric knowledge from existing knowledge graph (i.e., ConceptNet) and parametric knowledge from large language model (i.e., COMET). }
    \vskip -0.8em
    \label{fig:venn_diagram}
\end{figure}
We compare the extracted visual knowledge with the non-parametric knowledge in the existing knowledge graph (KG) and the parametric knowledge from the large language model (LLM). The comparison insights from the three knowledge resources are shown in the Venn Diagram in Figure~\ref{fig:venn_diagram}. 

\textbf{Compare with non-parametric knowledge.} We take ConceptNet~\cite{speer2017conceptnet} as the representative in the comparison with non-parametric knowledge. To map the knowledge generated by \ours to ConceptNet, we parse the knowledge into triplets and associate the endpoints of these triplets with nodes in ConceptNet. 
Then we calculate the similarity of embeddings\footnote{Embeddings are produced by ConceptNet API: https://github.com/commonsense/conceptnet-numberbatch.} between the parsed relation and all the edge relations among the mapped nodes in ConceptNet. If the similarity score exceeds a predetermined threshold, i.e., 0.75, we consider the mapping successful. 
As illustrated in Figure~\ref{fig:venn_diagram}, we observe that compared with the non-parametric knowledge in KG, the extracted visual knowledge captures richer and more meaningful spatial details, e.g., ``\textit{three layer cake on table}'', and motion dynamics, e.g., ``\textit{baby elephants walking around adventurous wood}''. 

% more locational information (e.g. plant-related or outside field-related or food-like things with locational information) than KG. What's more, the extracted visual knowledge demonstrates other kinds of examples: something has something or someone doing something. 

\textbf{Compare with parametric knowledge.} We compare with parametric knowledge contained in LLM by prompting the gpt-3.5-turbo\footnote{https://platform.openai.com/docs/models/gpt-3-5} model with the object information in the image. The prompt template used is detailed in Appendix~\ref{app:gpt}. The mapping process follows the approach mentioned earlier. It is found that compared with the parametric knowledge in LLM, the extracted visual knowledge exhibits unique fine-grained visual details, e.g., ``\textit{red sticker on fence}'', and provides precise scene information, e.g., ``\textit{the light shining from bright black background}''. 
% \hejie{prompting template and examples of generated parametric knowledge from LLM}
% prompting engineering

%%%%%%%%%%%%%%%%%%%%%%%%%%%%%%%%%%%%%%%%%%%%%%%%%%%%%%%%%%%%%%%%%%%%%%%%%%%%
\subsection{Ablation Study}
\begin{wrapfigure}{r}{0.6\textwidth}
\vspace{-4ex}
\includegraphics[width=\linewidth]{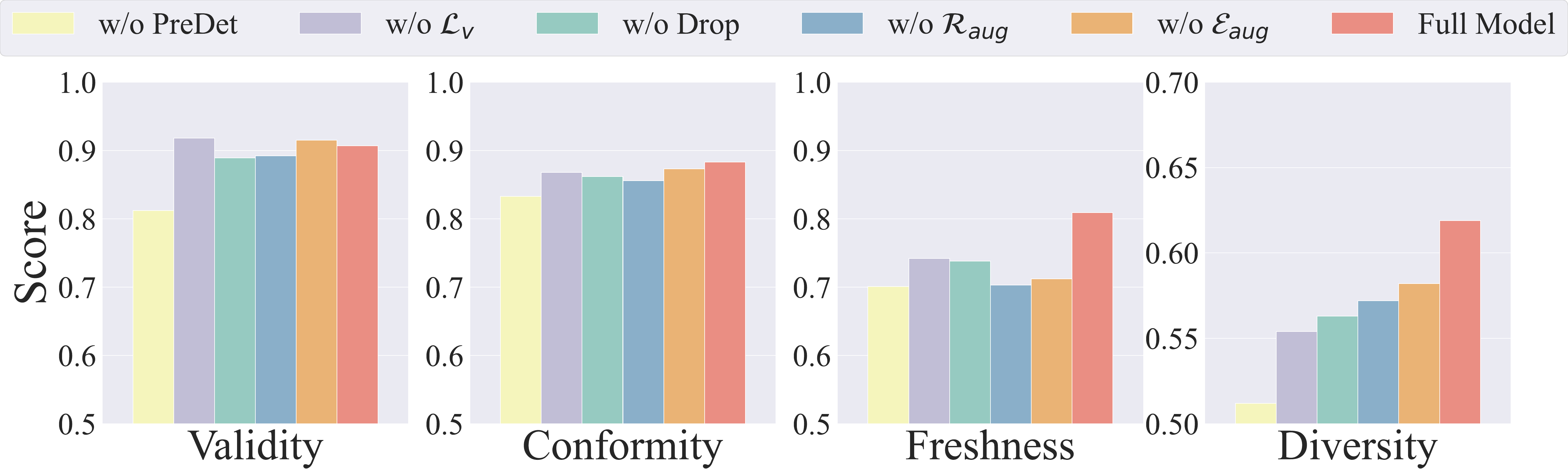} 
\vskip -0.5em
\caption{The influence of information variety regularization and diversity-driven data enhancement strategies.}
\vskip -1.0em
\label{fig:ablation}
\end{wrapfigure}

% \begin{figure}[h]
%     \centering
%     % \includegraphics[width=0.5\linewidth]{figures/ablation.png}
%     \includegraphics[width=1.0\linewidth]{figures/ablation.pdf}
%     \caption{The influence of the diversity-driven data enhancement strategies.}
%     \label{fig:ablation}
% \end{figure}

\textbf{The influence on knowledge quality with information variety regularization and data strategies.}
We conducted ablation studies to evaluate the effectiveness of the information variety regularizer, $\mathcal{L}_\text{V}$, and our diversity-driven data enhancement strategies. This involves an in-depth assessment of knowledge quality on the same evaluation subset. The results are presented in Figure~\ref{fig:ablation}. It is evident from the results that our proposed information variety design primarily impacts freshness and diversity, without compromising validity and conformity.
For the freshness, the omission of data augmentation for entities and relations results in the most significant performance degradation. This implies the crucial role these strategies play in infusing novel knowledge into the generation process. As for diversity, the most notable changes in metrics are observed when the $\mathcal{L}_\text{V}$ and random dropping are removed. The strategy for augmenting entities and relations also plays a valuable role in enriching diversity.

\textbf{Ablation of the pre-training for the open relational region detector.}
We conducted a comparison of the outcomes when loading a pre-trained detector backbone versus training the detector from scratch, as shown by the yellow bar in Figure~\ref{fig:ablation}. Results demonstrate a noticeable decrease in both knowledge diversity and freshness, which indicates the importance of loading the pre-trained model for region detection. This may be because omitting the pre-training step of the FasterRCNN model tends to result in the detection of more overlapping regions, which in turn causes the drop.

\textbf{The influence on dataset diversity with data strategies.}
% \begin{wraptable}{r}{0.8\textwidth}
% \centering
% \resizebox{\linewidth}{!}{%
% \begin{tabular}{l|ccc|c}
% \toprule
% \multirow{2.5}{*}{\bf Knowledge Metrics}  & \multicolumn{3}{c|}{\bf Training Dataset} & \bf Generate Knowledge \\
% % \cmidrule(lr){2-4} \cmidrule(lr){5} 
% & \it{Visual Genome}~\cite{krishna2017visual} & \it{Relational Caps}~\cite{kim2019dense}  & \cellcolor{yellow}\it{Diversity Enhanced (Ours)} & \cellcolor{gray!15}\it{\ours (Ours)} \\
% \midrule
% \bf Diversity & 0.589 & 0.604 & \cellcolor{yellow} 0.632 &  \cellcolor{gray!15} 0.619 \\
% \bottomrule
% \end{tabular}
% }
% \caption{Diversity of existing and enhanced datasets and generated knowledge from \ours.}
% \label{tab:abl-data}
% \end{wraptable}
We conduct a direct analysis of the knowledge diversity of the existing datasets and our diversity-enhanced one, compared with the visual knowledge generated from \ours. The findings, presented in Table \ref{tab:abl-data}, show that the diversity-driven data enhancement strategies significantly boost knowledge diversity. Trained with this enhanced data, \ours can extract visual knowledge that exhibits greater diversity than that found in the \textit{Visual Genome} and \textit{Relational Caps}, indicating the advantage of \ours to format-free visual knowledge generation and its ability to yield richer knowledge diversity.

\begin{table}[h]
\centering
\vskip -0.4em
\caption{Diversity of existing and enhanced datasets and generated knowledge from \ours.}
\vskip -0.4em 
\resizebox{0.8\linewidth}{!}{%
\begin{tabular}{l|ccc|c}
\toprule
\multirow{2.5}{*}{\bf Metrics}  & \multicolumn{3}{c|}{\bf Training Dataset} & \bf Generate Knowledge \\
% \cmidrule(lr){2-4} \cmidrule(lr){5} 
& \it{Visual Genome}~\cite{krishna2017visual} & \it{Relational Caps}~\cite{kim2019dense}  & \cellcolor{yellow}\it{Diversity Enhanced (Ours)} & \cellcolor{gray!15}\ours \it{(Ours)} \\
\midrule
\bf Diversity & 0.589 & 0.604 & \cellcolor{yellow} 0.632 &  \cellcolor{gray!15} 0.619 \\
\bottomrule
\end{tabular}
}
\vskip -0.8em
\label{tab:abl-data}
\end{table}

%%%%%%%%%%%%%%%%%%%%%%%%%%%%%%%%%%%%%%%%%%%%%%%%%%%%%%%%%%%%%%%%%%%%%%%%%%%%
\subsection{Case Study}
We present two case studies in Figure~\ref{fig:case_study} (See Appendix \ref{app:quali_ffknowledge} for more) to showcase the format-free visual knowledge generated by \ours, in comparison to Visual Genome (Scene Graph and Region Description) and Relational Caps. 
Contrary to the rigidity of scene graphs, which strictly adhere to a predefined format, \ours can generate knowledge with a flexible semantic structure, not strictly bound to the sub-verb-obj format (e.g., ``\textit{blue post attached to wall with white letter}''). Examples of this adaptability are highlighted in red.
When compared to dense region descriptions, the relational knowledge extracted by \ours offers a deeper understanding of the multiple entity interactions within an image.
In comparison to Relational Caps, which mainly focus on interactions between two objects, \ours significantly broadens the diversity of relation with vivid verbs (e.g., ``\textit{attached to}'', ``\textit{adorning}''). Moreover, it introduces novel entities (e.g., ``\textit{post}'', ``\textit{mane}'') and enriches the knowledge representation with nuanced details (e.g., ``\textit{full of}'', ``\textit{striped}'') that are missed by Relational Caps.

\begin{figure}[h]
    \vspace{-0.8em}
    \centering
    \includegraphics[width=1.0\linewidth]{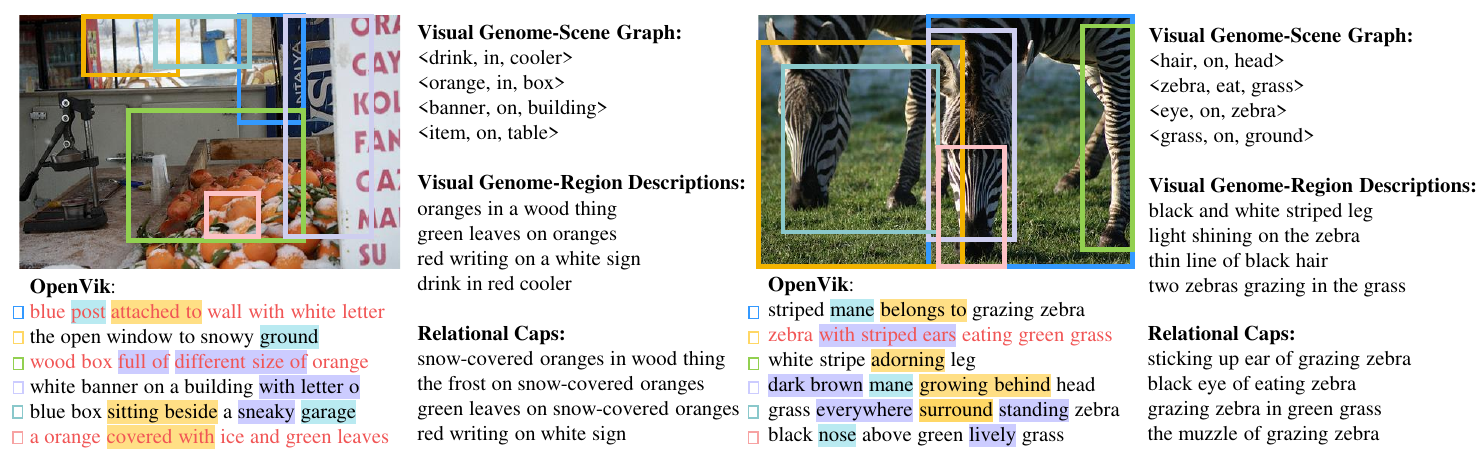}
    \caption{Case study on the extracted open visual knowledge from \ours. Examples of format-free knowledge are highlighted in {\color{case_ff}red}. Compared with VG and Relational Caps, \ours performs better at capturing novel {\colorbox{case_noun}{entities}}, broadening object interactions with diverse {\colorbox{case_verb}{relations}}, and enriching the knowledge representation with nuanced {\colorbox{case_adj}{descriptive details}}.
    % \manling{the descriptions in black color is mis-ordered. should switch the order for two examples}
    % \manling{need to explain the meaning of colors. it seems to be matched with Figure 2, but they are too far.}
    }
    \vskip -0.8em
    \label{fig:case_study}
\end{figure}

\begin{figure}[h!]
    \centering
    \includegraphics[width=\linewidth]{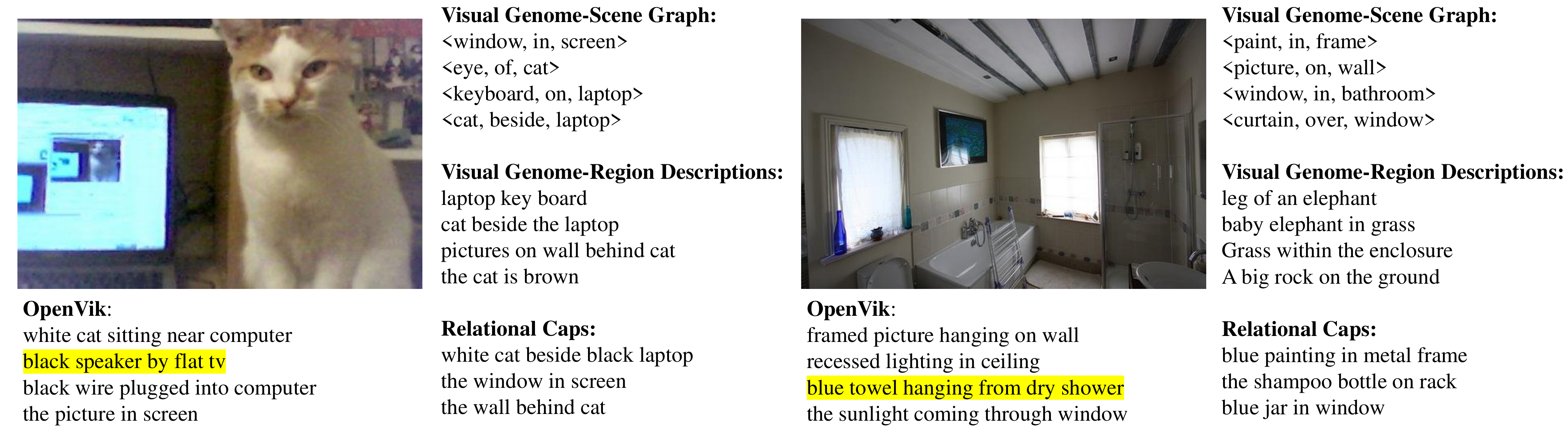}
    \caption{Examples of incorrectly knowledge resulting from distribution bias are {\colorbox{case_error}{highlighted}}.}
    \label{fig:failure_case}
\end{figure}

Note that we observe the unbalanced and noisy distributions within the training data can lead to errors in the knowledge produced. Viewing hallucinations as erroneous inferences based on input, the inaccuracies observed in OpenVik and similar baselines often stem from detection errors. These errors are typically caused by data biases that incorrectly associate features with a specific class or label. We further two illustrative failure cases in Figure~\ref{fig:failure_case}. For example, a ``\textit{black speaker by flat tv}'' is generated, although the speaker is not present in the image—possibly reflecting common co-occurrences within the dataset. Similarly, a ladder in the right figure has been misidentified as a towel, leading to the erroneous description of a ``\textit{blue towel hanging from dry shower}''. The key to mitigating such incorrect inference is identifying the cofounder feature of class labeling. 

% \begin{table}[h]
% \centering
% \small 
% \caption{
% Case Study on the knowledge generated by \ours and other datasets.
% }
% \resizebox{\textwidth}{!}{
% \begin{tabular}{lll}
% \toprule
% \textbf{Image}& \textbf{Dataset} & \textbf{Knowledge Examples} \\
% \midrule
% \multirow{11}{*}{\includegraphics[width=1cm, valign=c]{figures/1159413.jpg}} & 
% \multirow{2}{*}{\blue{xxx}} & {blalalalallalalallalalallal; bllalalalalallala; ballalalalallal; ballalalallalallallala; balallala} \\
% & & {xxx} \\ \\
% & \multirow{2}{*}{\orange{xxx}} & {xxx} \\
% & & {xxx} \\ \\
% & \multirow{2}{*}{\green{xxx}} & {xxx} \\
% & & {xxx} \\ \\
% & \multirow{2}{*}{\red{xxx}} & {xxx} \\
% & & {xxx.} \\
% \midrule
% \multirow{12}{*}{\textbf{2}} & 
% \multirow{2}{*}{\blue{xxx}} & {} \\
% & & {} \\
% \\
% & \multirow{2}{*}{\orange{xxx}} & {} \\
% & & {xxx} \\ 
% \\
% & \multirow{3}{*}{\green{xxx}} & {xxx} \\
% & & {xxx}  \\
% & & xxx \\
% \\
% & \multirow{2}{*}{\red{xxx}} & {xxx} \\
% & & {xxx} \\
% \bottomrule
% \end{tabular}
% }
% \label{tab:case_study}
% \end{table}
% \subsection{RQ3: Knowledge Utility Evaluation}
% \vspace{-0.3cm}
\section{Application}
\label{sec:application}
\vskip -0.6em

% observe the pattern (which kinds of samples are easier to be corrected by enriching the context from the visual knowledge graph) and take a subset if possible 
% 50% success, 50% failed (x)
This section explores whether the extracted open visual knowledge from \ours can bolster reasoning and inference capabilities in multimodality downstream tasks by augmenting a baseline in the challenging zero-shot setting.

%%%%%%%%%%%%%%%%%%%%%%%%%%%%%%%%%%
\subsection{Text-to-Image Retrieval}

% \noindent  $\diamond$ \textbf{Task Setting.} 
\textbf{Task Setting.} 
In the text-to-image retrieval task, a given caption is matched to a large set of candidate images, with the most relevant image returned as the result. Adopting the challenging zero-shot setting, we generate the visual representation $\bm v$ and textual representation $\bm t$ of the given image $\mathcal{I}$ and caption $\mathcal{T}$ using a pre-trained clip-retrieval model~\cite{beaumont-2022-clip-retrieval}. The baseline involves the image and text embedding similarly based on zero-shot CLIP Retrieval~\cite{beaumont-2022-clip-retrieval} and the fine-tuned model from BLIP~\cite{li2023blip}.

To explore the potential of the extracted visual knowledge from \ours, we enrich the given caption $\mathcal{T}$ with related contexts derived from the extracted visual knowledge. Specifically, for each query caption, we parse the caption to extract all subject-object pairs $(s, o)$ with the NLTK parser.
% \xin{how to conduct the ``parse'', rule-based, or procedure methods?}
Then $s$ and $o$ are mapped to the open visual knowledge, where knowledge phrases that contain relations occurring more than 30\% of the time between $s$ and $o$ are enriched to the original caption $\mathcal{T}$.

% \xinyu{Is it possible to reduce the task setting to a single paragraph}
\vspace{-0.2cm}
\begin{figure}[ht]
\begin{minipage}[c]{0.55\linewidth}
\centering
\includegraphics[width=\linewidth]{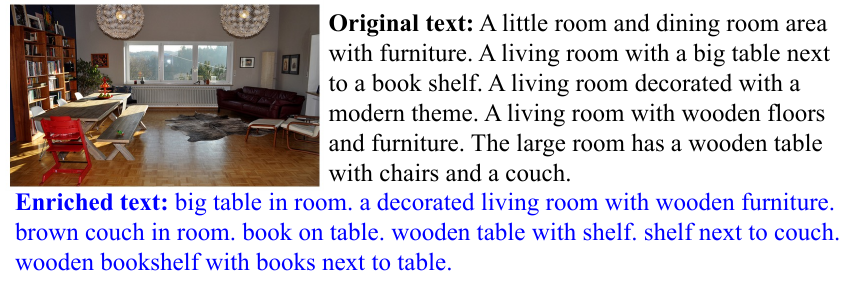}
\vskip -1.0em
\caption{An example of \ours enrichment on text-to-image retrieval (See Appendix \ref{app:quali_utility_ir} for more).}
\label{fig:ir}
\end{minipage}
\hfill
\begin{minipage}[c]{0.42\linewidth}
\vskip -2.4em
\centering
\resizebox{\linewidth}{!}{%
\begin{tabular}{l|ccccc}
\toprule
\textbf{Method} & \textbf{Recall@1} & \textbf{Recall@5}  & \textbf{Recall@10} & \textbf{\textit{Avg}}   \\
\midrule
% Zero-Shot & 0.36156351791530944 & 0.6547231270358306 & 0.7866449511400652 & 0.6009771986970684 &  \\
% \ours Context+ & 0.40553745928338764 & 0.7328990228013029 & 0.8452768729641694 & 0.6612377850162866 & \\
ZS-CLIP & 36.16 & 65.47 & 78.66 & 60.10 &  \\
\rowcolor{gray!15}\ours + ZS-CLIP  & 40.55 & 73.29 & 84.53 & 66.12 & \\
BLIP & 63.11 & 86.30 & 91.10 & 80.17 &  \\
\rowcolor{gray!15}\ours + BLIP  & 65.23 & 87.71 & 91.90 & 81.61 &  \\
\bottomrule

\end{tabular}
}
\captionof{table}{Text-to-image retrieval results (\%) of \ours enrichment compared with zero-shot baselines.}
% \caption{Quantitative results (\%) of \ours context enrichment on the text-to-image retrieval task compared with zero-shot baseline. %\xin{using \textit{Avg} would be better.}
% }
\vskip -1.0em
\label{tab:ir}
\end{minipage}
\end{figure}
% \vskip -1.0em

% \noindent  $\diamond$ \textbf{Qualitative examples.} 
\textbf{Qualitative examples.} 
% \begin{figure*}[h]
% \vskip -0.5em
%     \centering
%     \includegraphics[width=\linewidth]{figures/ir.pdf}
%     \caption{Examples of the context-enrichment corrected cases on the task of image retrieval.}
%     \vskip -0.5em
%     \label{fig:ir}
% \end{figure*}
% \begin{wrapfigure}{l}{0.55\textwidth}
% \vspace{-4ex}
% \includegraphics[width=\linewidth]{figures/ir.pdf} 
% \vskip -0.5em
% \caption{Examples of the context-enrichment corrected cases on the task of image retrieval.}
% \vskip -1.0em
% \label{fig:ir}
% \end{wrapfigure}
Figure~\ref{fig:ir} presents an example of \ours-based visual knowledge enrichment on captions. By incorporating related contexts from the generated open visual knowledge, the enriched captions convey more precise visual details, which enhances the alignment for text-image alignment.

% \noindent  $\diamond$ \textbf{Quantitative results.} 
\textbf{Quantitative results.} 
% \begin{table}[h]
% \centering
% \caption{Quantitative results (\%) on text-to-image retrieval before and after visual context enrichment.}
% \resizebox{0.65\linewidth}{!}{%
% \begin{tabular}{l|ccccc}
% \toprule
% \textbf{Method} & \textbf{Recall@1} & \textbf{Recall@5}  & \textbf{Recall@10} & \textbf{Recall@AVG}   \\
% \midrule
% % Zero-Shot & 0.36156351791530944 & 0.6547231270358306 & 0.7866449511400652 & 0.6009771986970684 &  \\
% % \ours Context+ & 0.40553745928338764 & 0.7328990228013029 & 0.8452768729641694 & 0.6612377850162866 & \\
% Zero-Shot & 36.16 & 65.47 & 78.66 & 60.10 &  \\
% \rowcolor{gray!15}\ours Context+ & 40.55 & 73.29 & 84.53 & 66.12 & \\
% \bottomrule
% \end{tabular}
% }
% \label{tab:ir}
% \end{table}
% \begin{wraptable}{r}{0.55\textwidth}
% \centering
% \vspace{-4.4ex}
% \caption{Quantitative results (\%) on text-to-image retrieval before and after visual context enrichment.}
% \resizebox{\linewidth}{!}{%
% \begin{tabular}{l|ccccc}
% \toprule
% \textbf{Method} & \textbf{Recall@1} & \textbf{Recall@5}  & \textbf{Recall@10} & \textbf{Recall@AVG}   \\
% \midrule
% % Zero-Shot & 0.36156351791530944 & 0.6547231270358306 & 0.7866449511400652 & 0.6009771986970684 &  \\
% % \ours Context+ & 0.40553745928338764 & 0.7328990228013029 & 0.8452768729641694 & 0.6612377850162866 & \\
% Zero-Shot & 36.16 & 65.47 & 78.66 & 60.10 &  \\
% \rowcolor{gray!15}\ours Context+ & 40.55 & 73.29 & 84.53 & 66.12 & \\
% \bottomrule
% \end{tabular}
% }
% \vskip -0.8em
% \label{tab:ir}
% \end{wraptable}
We curated a subset of 680 images from the testing set of the MS-COCO dataset containing parsed knowledge with at least eight nouns. This ensures an adequate degree of enrichment is achieved through the use of \ours. Standard image retrieval metrics, i.e., \textit{Recall@1/5/10/} and \textit{Avg}, are employed to evaluate the performance. The results are presented in Table~\ref{tab:ir}. It is evident that relational context enrichment leads to the average correction of more than 6.0\% of the initial zero-shot, highlighting the practical benefits of extracted visual knowledge in visual reasoning tasks.
%\xin{is there any other baseline with the same backbone but different knowledge from other models? Or any way to emphasize the \textbf{relational} knowledge instead of entity knowledge.}

%%%%%%%%%%%%%%%%%%%%%%%%%%%%%%%%%%
\subsection{Grounded Situation Recognition}
% \noindent  $\diamond$ \textbf{Task setting.} 
\textbf{Task setting.} 
% In the event type prediction of grounded situation recognition task, given an image $\mathcal{I}$, the objective is to predict an event type (e.g., ``eating'') by selecting from 504 event candidates~\cite{pratt2020grounded}.
The event type prediction for the grounded situation recognition task is to predict the best match from predefined 504 event types~\cite{pratt2020grounded} based on the image. We convert each candidate event verb into a description $\mathcal{T}$: ``\textit{An image of <verb>}'' for image description matching. Similarly to text-to-image retrieval, we include zero-shot CLIP and the fine-tuned model from BLIP as baselines. 

To enrich with contextual knowledge from \ours, for each given verb $v$, we find its nearest synonym in the extracted open visual knowledge and enrich the text description with the most common knowledge phrase containing it, regularized by the objects present in the image. Instead of directly concatenating the retrieved knowledge triplets to the original textual description, we employ an additive decomposition strategy: the similarity $s(\mathcal{I}, v)$ of the candidate verb $v$ with respect to the given image $\mathcal{I}$ is calculated as
$
s(\mathcal{I}, v)=\frac{1}{|D(v)|} \sum_{d \in D(v)} \phi(\mathcal{I}, v),
$
where $D(v)$ is the set of descriptors, including the original description and the enriched ones, and $\phi$ represents the single $\log$ probability that descriptor $d$ pertains to the image $\mathcal{I}$.

\begin{figure}[ht]
\begin{minipage}[c]{0.55\linewidth}
\centering
\includegraphics[width=\linewidth]{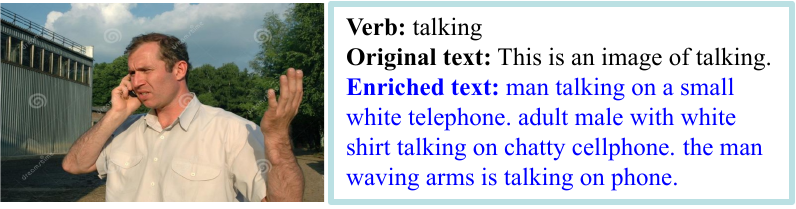}
\vskip -0.8em
\caption{An example of \ours context enrichment on task GSR (See Appendix \ref{app:quali_utility_gsr} for more).}
\label{fig:gsr}
\end{minipage}
\hfill
\begin{minipage}[c]{0.4\linewidth}
\vskip -1.4em
\centering
\resizebox{\linewidth}{!}{%
\begin{tabular}{l|ccccc}
\toprule
\textbf{Method} & \textbf{Accuracy} & \textbf{Precision}  & \textbf{Recall} & \textbf{$\text{F}_1$}   \\
\midrule
ZS-CLIP & 53.14 & 42.54 & 45.19 & 43.82 &  \\
\rowcolor{gray!15}\ours + ZS-CLIP & 75.16 & 61.63 & 62.75 & 62.18 &  \\
BLIP & 70.42 & 65.32 & 69.25 & 67.23 &  \\
\rowcolor{gray!15}\ours + BLIP & 80.25 & 72.55 & 70.61 & 71.57 & \\
% Zero-Shot & 0.5314465408805031 & 0.4253875968992248 & 0.45187338501291996 & 0.4178311352729957 &  \\
% OViK Context+ & 0.7515723270440252 & 0.6163237311385459 & 0.6275034293552811 & 0.6103141942648116 &  \\
\bottomrule
\end{tabular}
}
\captionof{table}{Grounded situation recognition results (\%) of \ours enrichment compared with zero-shot baselines.}
% \caption{Quantitative results (\%) of \ours context enrichment on GSR compared with zero-shot baseline.}
\vskip -1.0em
\label{tab:gsr}
\end{minipage}
\end{figure}

% \noindent  $\diamond$ \textbf{Qualitative examples.}
\textbf{Qualitative examples.}
% \begin{figure*}[h]
% % \vskip -1em
%     \centering
%     \includegraphics[width=0.95\linewidth]{figures/gsr.pdf}
%     \caption{Examples of the corrected cases with the visual knowledge enrichment on the task of grounded situation recognition.}
%     \vskip -0.5em
%     \label{fig:gsr}
% \end{figure*}
% \begin{wrapfigure}{l}{0.55\textwidth}
% \vspace{-4ex}
% \includegraphics[width=\linewidth]{figures/gsr.pdf} 
% \vskip -0.5em
% \caption{Examples of the corrected cases with the visual knowledge enrichment on the task of grounded situation recognition.}
% \vskip -1.0em
% \label{fig:gsr}
% \end{wrapfigure}
Figure~\ref{fig:gsr} presents a qualitative example of \ours-based context enrichment in the grounded situation recognition task. We observed that verbs like ``shopping'' and ``talking'' were appropriately enriched with their frequently occurring contexts from the open visual knowledge, leading to a reduced embedding distance between the description and its matching image.

% \noindent  $\diamond$ \textbf{Quantitative results.} 
\textbf{Quantitative results.} 
% \begin{table}[h]
% \centering
% \caption{Quantitative results (\%) on grounded situation recognition before and after visual context enrichment.}
% \resizebox{0.55\linewidth}{!}{%
% \begin{tabular}{l|ccccc}
% \toprule
% \textbf{Method} & \textbf{Accuracy} & \textbf{Precision}  & \textbf{Recall} & \textbf{$\text{F}_1$}   \\
% \midrule
% Zero-Shot & 53.14 & 42.54 & 45.19 & 41.78 &  \\
% \rowcolor{gray!15}\ours Context+ & 75.16 & 61.63 & 62.75 & 61.03 &  \\
% % Zero-Shot & 0.5314465408805031 & 0.4253875968992248 & 0.45187338501291996 & 0.4178311352729957 &  \\
% % OViK Context+ & 0.7515723270440252 & 0.6163237311385459 & 0.6275034293552811 & 0.6103141942648116 &  \\
% \bottomrule
% \end{tabular}
% }
% \label{tab:gsr}
% \end{table}
% \begin{wraptable}{r}{0.5\textwidth}
% \centering
% % \vskip -1.2em
% \vspace{-4.4ex}
% \caption{Quantitative results (\%) on grounded situation recognition before and after visual context enrichment.}
% % \vskip -0.5em 
% \resizebox{\linewidth}{!}{%
% \begin{tabular}{l|ccccc}
% \toprule
% \textbf{Method} & \textbf{Accuracy} & \textbf{Precision}  & \textbf{Recall} & \textbf{$\text{F}_1$}   \\
% \midrule
% Zero-Shot & 53.14 & 42.54 & 45.19 & 41.78 &  \\
% \rowcolor{gray!15}\ours Context+ & 75.16 & 61.63 & 62.75 & 61.03 &  \\
% % Zero-Shot & 0.5314465408805031 & 0.4253875968992248 & 0.45187338501291996 & 0.4178311352729957 &  \\
% % OViK Context+ & 0.7515723270440252 & 0.6163237311385459 & 0.6275034293552811 & 0.6103141942648116 &  \\
% \bottomrule
% \end{tabular}
% }
% \vskip -0.8em
% \label{tab:gsr}
% \end{wraptable}
We assembled a test set of 900 samples from the testing set of GSR that included verbs such as ``talking'', ``filming'', and ``picking'', among others, from a list of 256 words that can be accurately mapped to extracted visual knowledge, as well as 138 verbs that have a fuzzy match through ConceptNet embedding comparison. The full lists of the exact and fuzzy-matched verbs are detailed in Appendix \ref{app:verb_list}. The evaluated metrics include Accuracy, Precision, Recall, and $F_1$. The results are presented in Table~\ref{tab:gsr}. It can be observed that knowledge enrichment significantly outperforms the zero-shot and BLIP baselines. This suggests that the verb-related contexts introduced by \ours-generated knowledge are intuitive and greatly assist in understanding the semantics of event verbs, bolstered by related visual information.

%%%%%%%%%%%%%%%%%%%%%%%%%%%%%%%%%%
\subsection{Visual Commonsense Reasoning}
% \noindent  $\diamond$ \textbf{Task setting.} 
\textbf{Task setting.} 
The goal of visual commonsense reasoning is to predict an answer from four given option candidates for a given image and question. For the baseline approach, we compare the backbone model R2C from the VCR paper~\cite{zellers2019recognition} and BLIP~\cite{DBLP:conf/icml/0001LXH22}. In the visual knowledge-enhanced \ours Enriched approach, we perform two-level context augmentation, incorporating both entities and relations: (1) we parse the question and options to obtain all \verb|(S, O)| pairs and, for each entity pair, apply the same relation augmentation as in the image retrieval task; (2) for the \verb|V| in each option, we enrich the visual context using the same method as illustrated in grounded situation recognition.
\vskip -0.5em
\begin{figure}[ht]
\begin{minipage}[c]{0.57\linewidth}
\centering
\includegraphics[width=\linewidth]{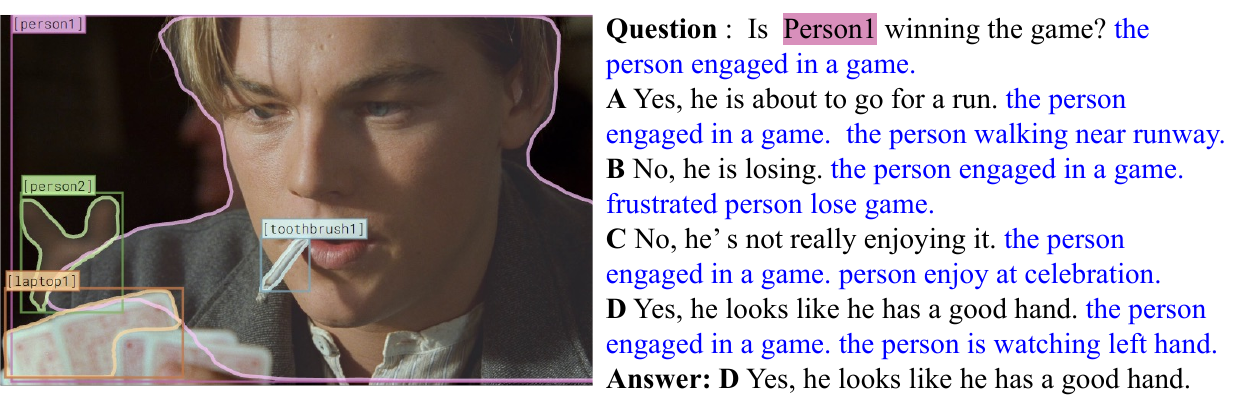}
\vskip -0.8em
\caption{An example of \ours context enrichment on the VCR task (See Appendix \ref{app:quali_utility_vcr} for more).}
\label{fig:vcr}
\end{minipage}
\hfill
\begin{minipage}[c]{0.4\linewidth}
\vskip -2.4em
\centering
\resizebox{\linewidth}{!}{%
\begin{tabular}{l|ccccc}
\toprule
\textbf{Method} & \textbf{Accuracy} & \textbf{Precision}  & \textbf{Recall} & \textbf{$\text{F}_1$}   \\
\midrule
R2C  & 56.66 & 56.73 & 56.72 & 56.72 &  \\
\rowcolor{gray!15}\ours + R2C & 59.96 & 60.01 & 60.03 & 60.02 &  \\
BLIP & 62.50 & 62.50 & 62.45 & 62.47 &  \\
\rowcolor{gray!15}\ours + BLIP & 67.40 & 67.54 & 67.43 & 67.48 &  \\
% Zero-Shot & 0.5665601703940362 & 0.5672774800934536 & 0.5672298526972793 & 0.5665692934357616 &  \\
% \ours Context+ & 0.5995740149094781 & 0.6000953382873933 & 0.6003232731335536 & 0.5994755968967521 &  \\
\bottomrule
\end{tabular}
}
\captionof{table}{Visual commonsense reasoning results (\%) of \ours context enrichment compared with zero-shot baselines.}
% \caption{Quantitative results (\%) of \ours context enrichment on the visual commonsense reasoning task compared with zero-shot baseline.}
\vskip -1.0em
\label{tab:vcr}
\end{minipage}
\end{figure}

% \noindent  $\diamond$ \textbf{Qualitative examples.}
\textbf{Qualitative examples.}
% \begin{figure*}[h]
% % \vskip -1em
%     \centering
%     \includegraphics[width=\linewidth]{figures/vcr.pdf}
%     \caption{Examples of the corrected cases with the visual context knowledge enrichment on the task of visual commonsense reasoning.}
%     % \vskip -1em
%     \label{fig:vcr}
% \end{figure*}
% \begin{wrapfigure}{l}{0.55\textwidth}
% \vspace{-4ex}
% \includegraphics[width=\linewidth]{figures/vcr.pdf} 
% \vskip -0.5em
% \caption{Examples of the corrected cases with the visual context knowledge enrichment on the task of visual commonsense reasoning.}
% \vskip -1.0em
% \label{fig:vcr}
% \end{wrapfigure}
Figure \ref{fig:vcr} presents an example before and after applying the two-level visual knowledge-based enrichment for visual commonsense reasoning. The results indicate that visual knowledge enhances the correspondence between the correct answer and the image itself. 

% \noindent  $\diamond$ \textbf{Quantitative results.} 
\textbf{Quantitative results.} 
% \begin{table}[h]
% \centering
% \caption{Quantitative results (\%) on visual commonsense reasoning before and after visual context enrichment.}
% \resizebox{0.55\linewidth}{!}{%
% \begin{tabular}{l|ccccc}
% \toprule
% \textbf{Method} & \textbf{Accuracy} & \textbf{Precision}  & \textbf{Recall} & \textbf{$\text{F}_1$}   \\
% \midrule
% Zero-Shot & 56.66 & 56.73 & 56.72 & 56.66 &  \\
% \rowcolor{gray!15}\ours Context+ & 59.96 & 60.01 & 60.03 & 59.95 &  \\
% % Zero-Shot & 0.5665601703940362 & 0.5672774800934536 & 0.5672298526972793 & 0.5665692934357616 &  \\
% % \ours Context+ & 0.5995740149094781 & 0.6000953382873933 & 0.6003232731335536 & 0.5994755968967521 &  \\
% \bottomrule
% \end{tabular}
% }
% \label{tab:vcr}
% \end{table}
% \begin{wraptable}{r}{0.5\textwidth}
% \centering
% \vspace{-4.4ex}
% \caption{Quantitative results (\%) on visual commonsense reasoning before and after visual context enrichment.}
% \resizebox{\linewidth}{!}{%
% \begin{tabular}{l|ccccc}
% \toprule
% \textbf{Method} & \textbf{Accuracy} & \textbf{Precision}  & \textbf{Recall} & \textbf{$\text{F}_1$}   \\
% \midrule
% Zero-Shot & 56.66 & 56.73 & 56.72 & 56.66 &  \\
% \rowcolor{gray!15}\ours Context+ & 59.96 & 60.01 & 60.03 & 59.95 &  \\
% % Zero-Shot & 0.5665601703940362 & 0.5672774800934536 & 0.5672298526972793 & 0.5665692934357616 &  \\
% % \ours Context+ & 0.5995740149094781 & 0.6000953382873933 & 0.6003232731335536 & 0.5994755968967521 &  \\
% \bottomrule
% \end{tabular}
% }
% \vskip -0.8em
% \label{tab:vcr}
% \end{wraptable}
We assembled a test set of 939 samples from the validation set of the VCR dataset~\cite{zellers2019recognition}. Each sample in this test set contains questions and answers with a minimum of five nouns and two relations, guaranteeing an adequate level of information complexity for meaningful engagement with open visual knowledge. The results can be found in Table~\ref{tab:vcr}. We observe that the enriched visual knowledge helps especially when solving reasoning questions on humans and their interactions with visually impressive entities, such as ``\textit{game}'' in Figure \ref{fig:vcr}. This enhancement results in a performance improvement above 3.0\% over the zero-shot baseline.

%%%%%%%%%%%%%%%%%%%%%%%%%%%%%%%%%%%%%%%%%%%%%%%%%%%%%%%%%%%%%%%%%%%%%%%%%%%%
% \subsection{Hyper-parameter Analysis}s
\section{Conclusion, Limitations, and Future Work}
\label{sec:conclusion}
% In conclusion, our novel framework for extracting structured visual knowledge advances the field of visual knowledge discovery by addressing the challenges of integrating visual and textual modalities and generating diverse knowledge representations. The region detector and knowledge generator modules, combined with the data enrichment and model regularization strategies, contribute to a more comprehensive understanding of complex concepts and context-aware language generation. The proposed approach demonstrates its effectiveness in various evaluations and downstream tasks, but there remains potential for further improvement and exploration. 
This work is the first exploration of a new paradigm of open visual knowledge extraction, which combines an open relational region detector to flexibly pinpoint relational regions and a format-free visual knowledge generator that generates visual knowledge by prompting a multimodality model conditioned on the region of interest. To further enhance the diversity of the generated knowledge, we explore two distinct data enhancement techniques. Extensive knowledge evaluations underscore the correctness and uniqueness of our extracted open visual knowledge, and the consistent improvements observed across various visual reasoning tasks highlight the real-world applicability of \ours.
% While our work presents a significant step towards this new paradigm of open visual knowledge extraction, several limitations present possible avenues for future work. 

While our approach has been shown effective in various scenarios, its performance at larger scales or on more diverse datasets remains to be studied. Future work could investigate its effectiveness across a broader range of tasks and contexts. Also, the current model requires fine-tuning for the visual knowledge extractor. 
%Developing a model that can generalize well without fine-tuning could be another interesting direction for future work.
% Moreover, the development of novel techniques for evaluating the quality of extracted visual knowledge and its impact on downstream tasks will provide valuable insights for the research community and facilitate continued progress in the field of visual knowledge discovery.
Developing a model that can generalize well with prompt tuning or demonstration augmentation could be another interesting direction for future work.
\section{Acknowledgments}
\label{sec:ack}
Carl Yang was supported by the National Institute Of Diabetes And Digestive And Kidney Diseases of the National Institutes of Health under Award Number K25DK135913.

\clearpage
\bibliographystyle{plain}
\bibliography{reference}
\clearpage
\appendix
%%%%%%%%%%%%%%%%%%%%%%%%%%%%%%%%%%%%%%%%%%%%%%%%%%%%%%%%%
\section{Details of Data Augmentation with External Knowledge Resources}
\label{app:data_aug}
\noindent\underline{\textit{\ding{52} Enhance Relation Recognition}}: 
% To enhance relation recognition, subject-object pairs that appears in an image are initially extracted from the original training dataset. Frequent pairs undergo semantic enhancement by leveraging ConceptNet. This involves treating the subject and object as nodes and retrieving edges with weights above a certain threshold to augment the relationship between them.
We enriched the relationships between objects parsed from the original knowledge descriptions by leveraging the external resource of ConceptNet. ConceptNet comprises commonly observed entities and their connections, where edge weights signify the reliability and frequency of these relationships. The typical value of edge weights in ConceptNet is 1. To prevent the redundancy of common information and to maintain the validity of the enriched relations, we categorized the relationships based on their weights. Relationships with weights less than 1 were deemed ``weak'' and those with a weight of 1 were labeled ``average''. We refrained from using these categories for relation enhancement. Instead, only relationships with weights greater than 1, indicative of high reliability, were employed for augmenting the relations.

\noindent\underline{\textit{\ding{52} Boost Entity Perception}}: 
On the entity side, we augment complement entities and descriptive information with two external knowledge resources. 
On one hand, for descriptions with a high TF-IDF+ score, we enrich related entities of the object from ConceptNet to create additional knowledge descriptions. The relatedness is based on the between-word relatedness score provided by ConceptNet and we take the threshold as 0.85.
On the other hand, we employ the Commonsense Transformers (COMET)~\cite{DBLP:conf/acl/BosselutRSMCC19} model to enrich related new objects and descriptive information. The COMET model is a language model designed to generate commonsense knowledge and understand causal relationships between descriptions. It is pretrained using the atomic dataset, which consists of structured, crowd-sourced knowledge about everyday events and their associated causes and effects. The COMET model can provide neighbor descriptions of the given input of nine different categories of relation. We take the {\ttfamily xAttr} and {\ttfamily oEffect} relation categories and augmented the COMET model by formulating the existing knowledge description texts as the input and choose the corresponding category branch during generation for enriching objects and descriptions respectively.
% \begin{itemize}[nosep,leftmargin=*]
% \item {\ttfamily oEffect}: The effect the event in description has on others or other things besides the certain person. 
% \item {\ttfamily oReact}: The reaction of others besides the certain person to the event in the description.
% \item {\ttfamily oWant}: What others besides certain person may
% want to do after the event in the description happened.
% \item {\ttfamily xAttr}: How something might be described given their part in the event.
% \item {\ttfamily xEffect}: The effect that the event in the description would have on certain person.
% \item {\ttfamily xIntent}: The reason why the certain person would cause the event in the description.
% \item {\ttfamily xNeed}: What certain person might need to do before the event in the description.
% \item {\ttfamily xReact}:The reaction that certain person would
% have to the event in the description.
% \item {\ttfamily xWant}: What the certain person may want to do after
% the event in the description.
% \end{itemize}

% Due to subject and object nature uncertainty, employing all relations for semantic enhancement supplementation may lead to increased semantic ambiguity or perverse meaning of statements. Consequently, only the {\ttfamily xAttr} relation category is used to introduce attribute descriptions.

\section{Dataset Information}
\label{app:data}
\begin{table}[h]
\centering
\caption{Dataset statistics.}
\resizebox{0.7\linewidth}{!}{%
\begin{tabular}{ccccc}
\toprule
\textbf{split} & \textbf{\#image} & \textbf{\#descriptor}  & \textbf{\#relation} & \textbf{\#subject \& object}   \\
\midrule
% Train & 75,456 & 1,041,381 & 28,556 & 225,270   \\
% Validation & 4,871 & 82,459 & 4,258 & 32,198  \\
% Test & 4,873 & 81,496 & 4,202 & 31,798   \\
Train & 75,456 & 832,351 & 30,241 & 302,735   \\
Validation & 4,871 & 64,137 & 5,164 & 34,177  \\
Test & 4,873 & 62,579 & 5,031 & 32,384   \\
\bottomrule
\end{tabular}
}
\label{tab:dataset}
\end{table}
The statistic information of our augmented dataset is summarized in Table \ref{tab:dataset}, where \textbf{split} specifies the dataset split, \textbf{\#image} indicates the number of images in the split, \textbf{\#descriptor} indicates the total number of relational descriptors of the images, \textbf{\#relation} is the total number of unique relations in the relational descriptors after deduplication, and \textbf{\#subject \& object} is the total number of subjects and objects contained in the description text. 

%%%%%%%%%%%%%%%%%%%%%%%%%%%%%%%%%%%%%%%%%%%%%%%%%%%%%%%%%
\section{Implementation Details}
\label{app:imple}
% \label{app:data}
\begin{minipage}{\textwidth}
\begin{minipage}[t]{0.4\textwidth}
\makeatletter\def\@captype{table}
\begin{tabular}{c|c}
\toprule
\textbf{Hyperparameter} & \textbf{Assignment} \\
\midrule
% backbone & ResNet50-FPN  \\
% Text Supervision Model & $\text{BLIP}_\text{base}$  \\
batch size & 4   \\
learning rate optimizer & Adam \\
Adam epsilon & 1e-8 \\
Adam initial learning rate & 1e-5 \\
learning rate scheduler & cosine scheduler \\
Adam decay weight & 0.05 \\
% imgae embedding size & 4096 \\
% train epoch & 20\\
\bottomrule
\end{tabular}
\captionof{table}{Hyperparameters for training open relational region detector.}
\label{tab:hyper4detector}
\end{minipage}\qquad\qquad\quad
\begin{minipage}[t]{0.4\textwidth}
\makeatletter\def\@captype{table}
\begin{tabular}{c|c}
\toprule
\textbf{Hyperparameter} & \textbf{Assignment} \\
\midrule
% Vision Transformer model & ViT-B  \\
% Decoder Model & $\text{BLIP}_\text{base}$  \\
batch size & 4   \\
learning rate optimizer & Adam \\
Adam epsilon & 1e-8 \\
Adam initial learning rate & 1e-5 \\
learning rate scheduler & cosine scheduler \\
Adam decay weight & 0.05 \\
% imgae cropping size & 384 \\
% training epoch & 20\\
 $\alpha$ & 0.7\\
 $\phi$ & 0.01\\
\bottomrule
\end{tabular}
\captionof{table}{Hyperparameters for training format-free visual knowledge generator.}
\label{tab:hyper4generator}
\end{minipage}
\end{minipage}
% \begin{table}[h]
% \centering

\textbf{Open relational region detector.} 
The visual feature extraction backbone is constructed upon a pre-trained ResNet50-FPN. The detector head incorporates a $\text{BLIP}_\text{base}$ equipped with the essential ViT-B/16 for text supervision, using multiple fully connected layers to derive region features. For each candidate region, we engage a regressor to conduct boundary regression on these features. The detector undergoes fine-tuning for 20 epochs using the relational region bounding box dataset and an Adam optimizer~\citep{loshchilov2019decoupled}. The hyperparameters for training are detailed in Table ~\ref{tab:hyper4detector}.

% and a weight decay of 0.9. The batch size is set as 4. The learning rate is warmed-up to 1e-3 and decayed linearly with a rate of 0.85. The hidden dimension of the image representation is set as 4096, and a single token is mapped to a vector interpretation consisting of 4096 elements after undergoing the embedding layer. The detailed hyperparameters are summarized in Table
% The model detects a maximum of 30 bounding boxes for each image with the highest confidence to avoid misleading noises.

\textbf{Format-free visual knowledge generator.} 
The format-free visual knowledge generator is initialized from $\text{BLIP}_\text{base}$, which incorporates the basic ViT-B/16. We fine-tune the generator model for 20 epochs using the same optimizer as the one employed for the region detector. Detailed hyperparameters for the visual knowledge generator can be found in Table ~\ref{tab:hyper4generator}.
\section{Human Evaluation Guidance and Interface}
\label{app:human_eval}
We perform the human evaluation on two of the four in-depth knowledge quality assessment metrics.  
We build an interface by referring to~\cite{welleck2019neural}, where raters are presented with a given image and the corresponding knowledge descriptions and are required to choose one from the multiple choice for two questions on whether the knowledge is valid to humans and whether the knowledge description depicts the image. The detailed scoring criteria for \textit{Validity} and \textit{Conformity} are provided below: 
\begin{itemize}[nosep,leftmargin=*]
    \item \textit{Validity} ($\uparrow$): \textit{whether the generated visual knowledge is valid to humans}.
        \begin{itemize}
            \item 0 (Invalid): The knowledge description does not conform to human cognition, rendering it unreliable or misleading to humans.
            \item 1 (Valid): The knowledge description is valid and accurately conforms to human cognition, providing reliable and meaningful knowledge to humans.
        \end{itemize}
    \item \textit{Conformity} ($\uparrow$): \textit{whether the generated knowledge faithfully depicts the scenarios in the images}.
        \begin{itemize}
            \item 0 (Inconsistent): The knowledge description does not faithfully depict the scenarios in the images, showing significant deviations or discrepancies, making it difficult for users to relate the textual information to the visual context.
            \item 1 (Partially Conforming): The knowledge description partially conforms to the scenarios in the images, but there might be minor inconsistencies or missing relevant details.
            \item 2 (Moderately Conforming): The knowledge description exhibits a moderate level of conformity with the scenarios in the images, capturing the key aspects and providing coherent descriptions.
            \item 3 (Highly Conforming): The knowledge description highly conforms to the scenarios in the images, accurately capturing the details and faithfully representing the visual context.
        \end{itemize}
    % \item \textit{Freshness} ($\uparrow$): the novelty of the knowledge, i.e., the proportion not present in the training set.
    % \item \textit{Diversity} ($\uparrow$): the language variance between a randomly sampled pair of knowledge pieces.
\end{itemize}

\begin{figure}[h]
\includegraphics[width=\linewidth]{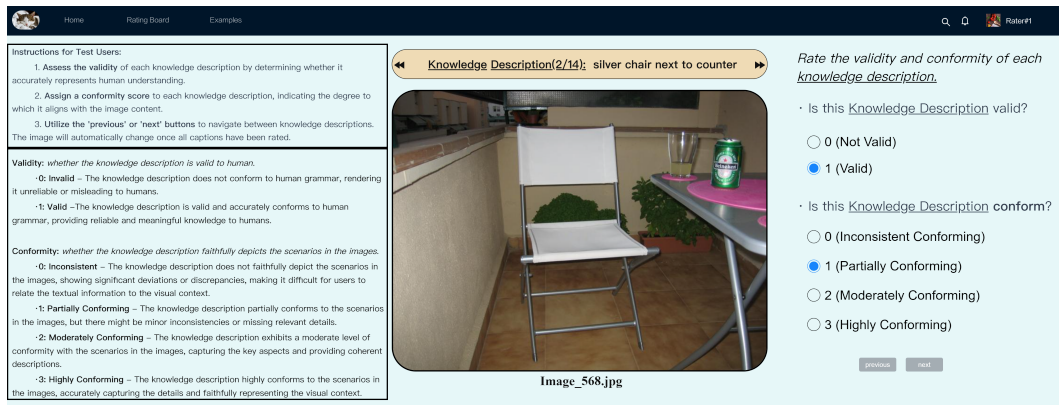}
\caption{The human evaluation interface for in-depth knowledge quality evaluation.}
\label{fig:interface}
\end{figure}

\textbf{Agreement/validation} 
We use Cohen's $\kappa$ as the agreement score to measure potential subjectivity involved in ratings of knowledge quality. Cohen's $\kappa$ is a statistic that is used to measure inter-rater reliability for qualitative items and is scaled from -1 (perfect systematic disagreement) to 1 (perfect agreement), where values $\le$ 0 as indicating \textit{no agreement} and 0.01-0.20 as \textit{none to slight}, 0.21-0.40 as \textit{fair}, 0.41–0.60 as \textit{moderate}, 0.61-0.80 as \textit{substantial}, and 0.81-1.00 as \textit{almost perfect} agreement. Our calculated average pairwise Cohen's $\kappa$ on human evaluation results from three different raters is 0.76, which indicates a good agreement.

%%%%%%%%%%%%%%%%%%%%%%%%%%%%%%%%%%%%%%%%%%%%%%%%%%%%%%%%%
\section{Parametric Knowledge Prompting Template}
\label{app:gpt}
Given an image $\mathcal{I}$ and the corresponding extracted visual knowledge from it based on \ours, we perform knowledge comparison with parametric knowledge contained in LLM by prompting the gpt-3.5-turbo model with the object information contained in the $\mathcal{I}$. The prompt format is shown in the followings: 
% \textit{"As a relational information extractor, use the object-subject pair information provided in the image to summarize it by extracting 5-10 condensed relational information using simple predicates and verbs. Provide the extracted information in a bullet list with diverse format."}
% \textit{"Assume you are relational information extractor. As a relational information extractor, use the object-subject pair information provided in the image to summarize it by extracting 5-10 condensed relational information using simple predicates and verbs. Provide the extracted information in a bullet list format with each item in the form of [object, relation, subject]."}.

\fbox{
  \begin{minipage}{13.5cm} % Adjust the width as needed
    % \texttt{As a relational information extractor, use the object-subject pair information provided in the image to summarize it by extracting 5-10 condensed relational information using simple predicates and verbs. Provide the extracted information in a bullet list with diverse formats.}
    \texttt{Suppose you are looking at an image that contains the following subject and object entities: \\
    Subject list: [Insert the subject names here] \\
    Object list: [Insert the object names here] \\
    Please extract 5-10 condensed descriptions that describe the interactions and/or relations among those entities in the image. Try to elucidate the associations and relationships with diverse language formats instead of being restricted to sub-verb-obj tuples.}
  \end{minipage}
}

%%%%%%%%%%%%%%%%%%%%%%%%%%%%%%%%%%%%%%%%%%%%%%%%%%%%%%%%%
\section{More Case Studies of Open Visual Knowledge from \ours}
\label{app:quali_ffknowledge}
\begin{figure}[h]
\includegraphics[width=\linewidth]{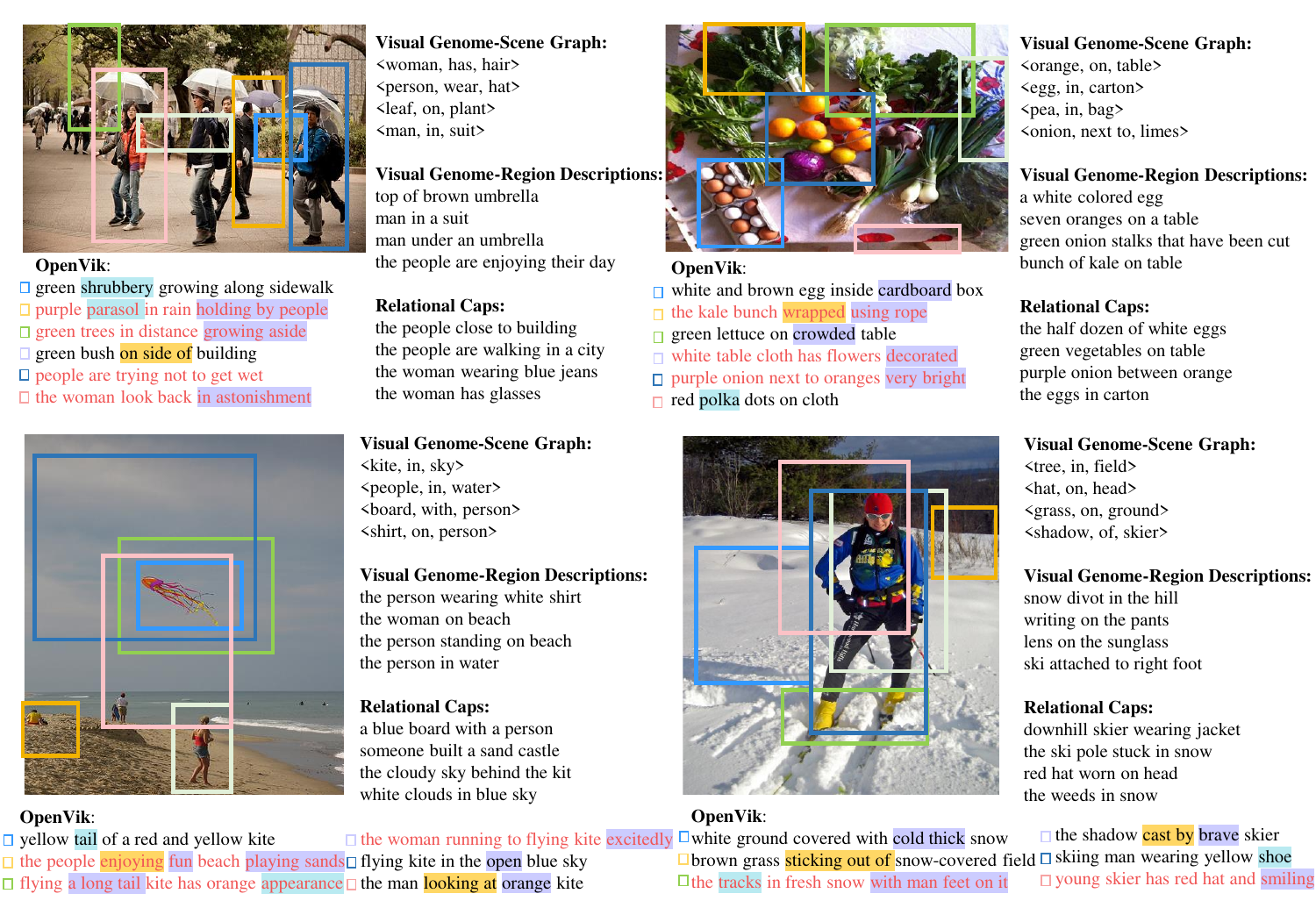}
\caption{Case studies of open visual knowledge from \ours.}
\label{fig:case-app-1}
\end{figure}

Figure ~\ref{fig:case-app-1} shows some other cases on the extracted open visual knowledge from \ours. 
In comparison to VG and Relational Caps, \ours exhibits superior performance at capturing novel {\colorbox{case_noun}{entities}}, expanding object interactions through diverse {\colorbox{case_verb}{relations}}, and enriching knowledge representation with nuanced {\colorbox{case_adj}{descriptive details}}. 
%Several different colored boxes are displayed to allow knowledge to be matched to regional information.\\
For example for the bottom right image, \ours can extract novel entities such as ``\colorbox{case_noun}{\textit{tracks}}'', ``\colorbox{case_noun}{\textit{shoe}}'', diverse relations such as ``\colorbox{case_verb}{\textit{sticking out of}}'', and nuanced descriptive details such as ``\colorbox{case_adj}{\textit{cold thick}}'', ``\colorbox{case_adj}{\textit{with man feet on it}}'', ``\colorbox{case_adj}{\textit{brave}}''. The generated knowledge with a more format-free semantic structure is highlighted in {\color{case_ff}red}. %It indicates that \ours can generate more format-free knowledge descriptions compared with other existing datasets. 
% for the image in the bottom right corner of figure ~\ref{fig:case-app-1}:
% Compared with VG and Relational Caps, \ours performs better at capturing novel {\colorbox{case_noun}{entities}}, broadening object interactions with diverse {\colorbox{case_verb}{relations}}, and enriching the knowledge representation with nuanced {\colorbox{case_adj}{descriptive details}}.

%%%%%%%%%%%%%%%%%%%%%%%%%%%%%%%%%%%%%%%%%%%%%%%%%%%%%%%%%
\section{More Qualitative Examples on Applications}
\label{app:quali_utility}
\subsection{Text-to-Image Retrieval}
\label{app:quali_utility_ir}
\begin{figure}[h]
\includegraphics[width=\linewidth]{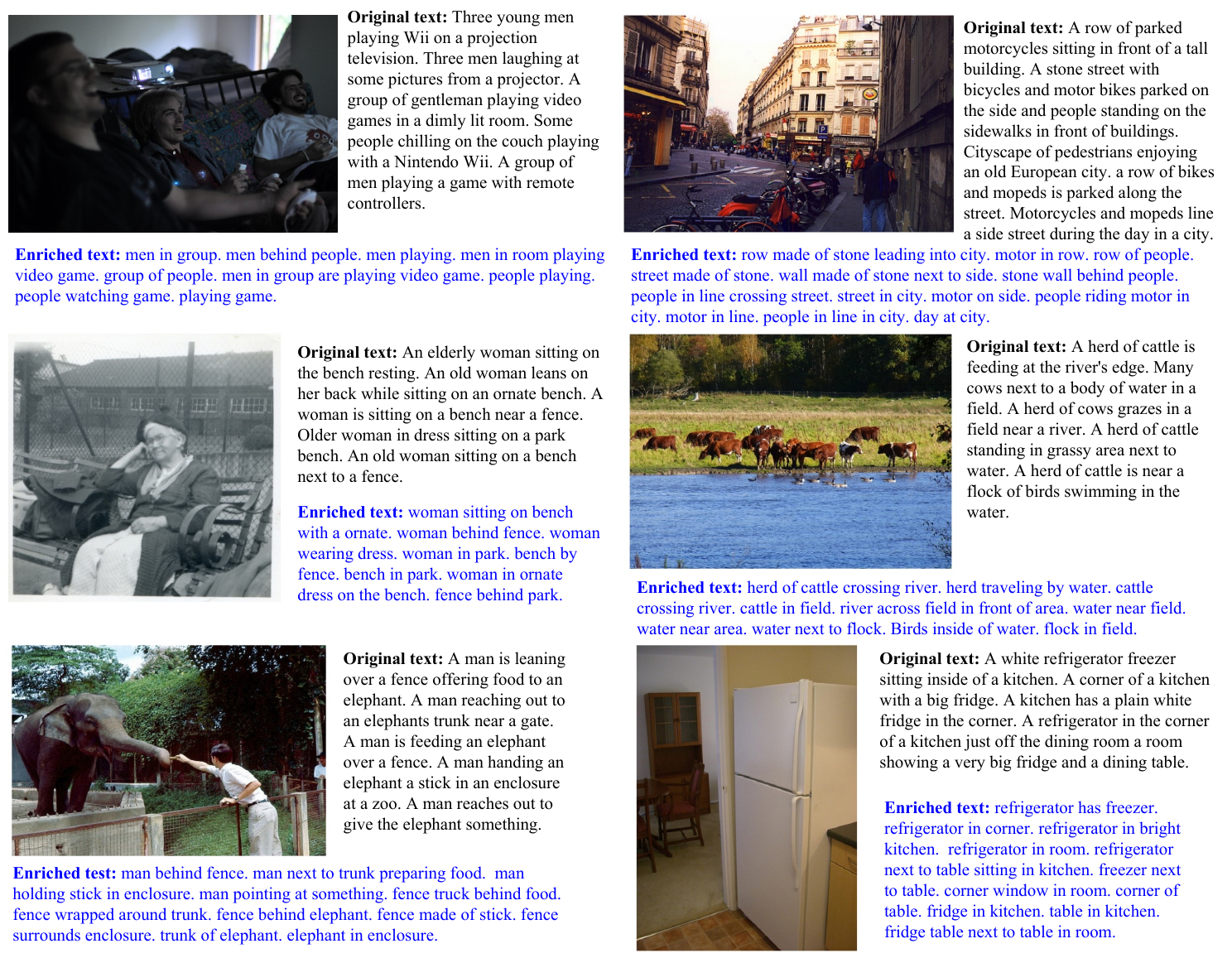}
\caption{Qualitative examples of \ours context enrichment on text-to-image retrieval.}
\label{fig:irapp}
\end{figure}
Figure~\ref{fig:irapp} presents more qualitative examples of \ours-based visual knowledge enrichment on captions. The enriched text is based on the objects present in the images themselves, supplemented with additional relationships from our generated visual knowledge in \ours. It is shown that the introduced relationships often provide new context information that aligns with the visual content of the images. For example, in the image of an old woman sitting on a bench in a park, the enriched context information includes the positional relationship between the ``\textit{bench}'', ``\textit{fence}'', and ``\textit{park}'', which provides a more comprehensive description of the original image.

\subsection{Grounded Situation Recognition}
\label{app:quali_utility_gsr}
\begin{figure}[h]
\includegraphics[width=\linewidth]{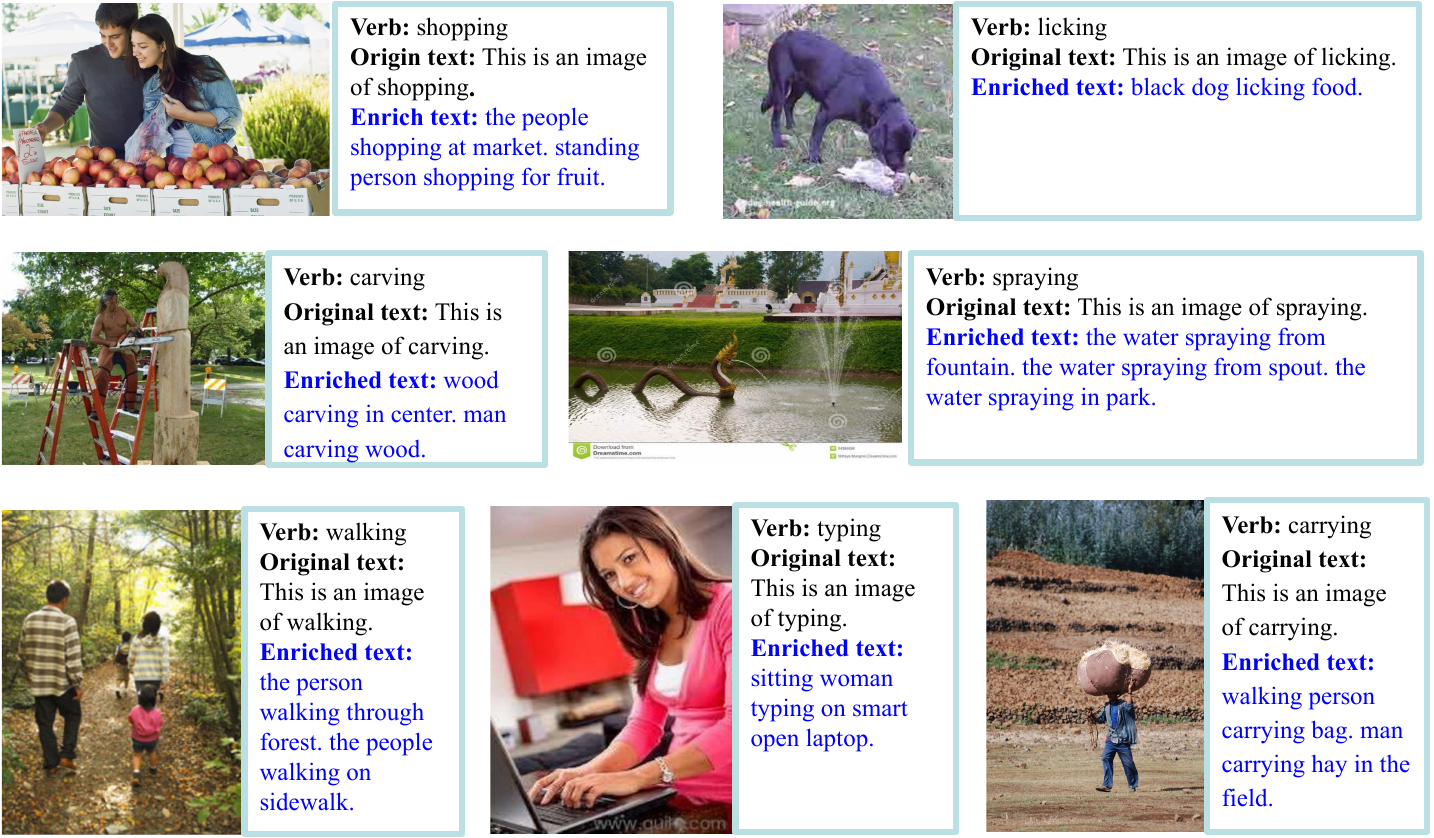}
\caption{Qualitative examples of \ours context enrichment on task GSR.}
\label{fig:gsrapp}
\end{figure}
Figure~\ref{fig:gsrapp} presents more qualitative examples of \ours-based context enrichment in the grounded situation recognition (GSR) task.
Our context enrichment setting for the GSR task is to perform enrichment based on verbs like ``\textit{shopping}'' and ``\textit{carrying}''. We further restrict the enriched context with the objects contained in the image to avoid noisy enrichment. For example, for the image showing people shopping at a market, the enriched knowledge contexts could be ``\textit{the people shopping at market}'', ``\textit{standing person shopping for fruit}''. The idea is to enrich the original description $\mathcal{T}$: ``\textit{An image of <verb>}'' with relevant actions and relations with the extracted visual knowledge from \ours, which can potentially help in drawing-in the matched candidates.

\subsection{Visual Commonsense Reasoning}
\label{app:quali_utility_vcr}
\begin{figure}[h]
\includegraphics[width=\linewidth]{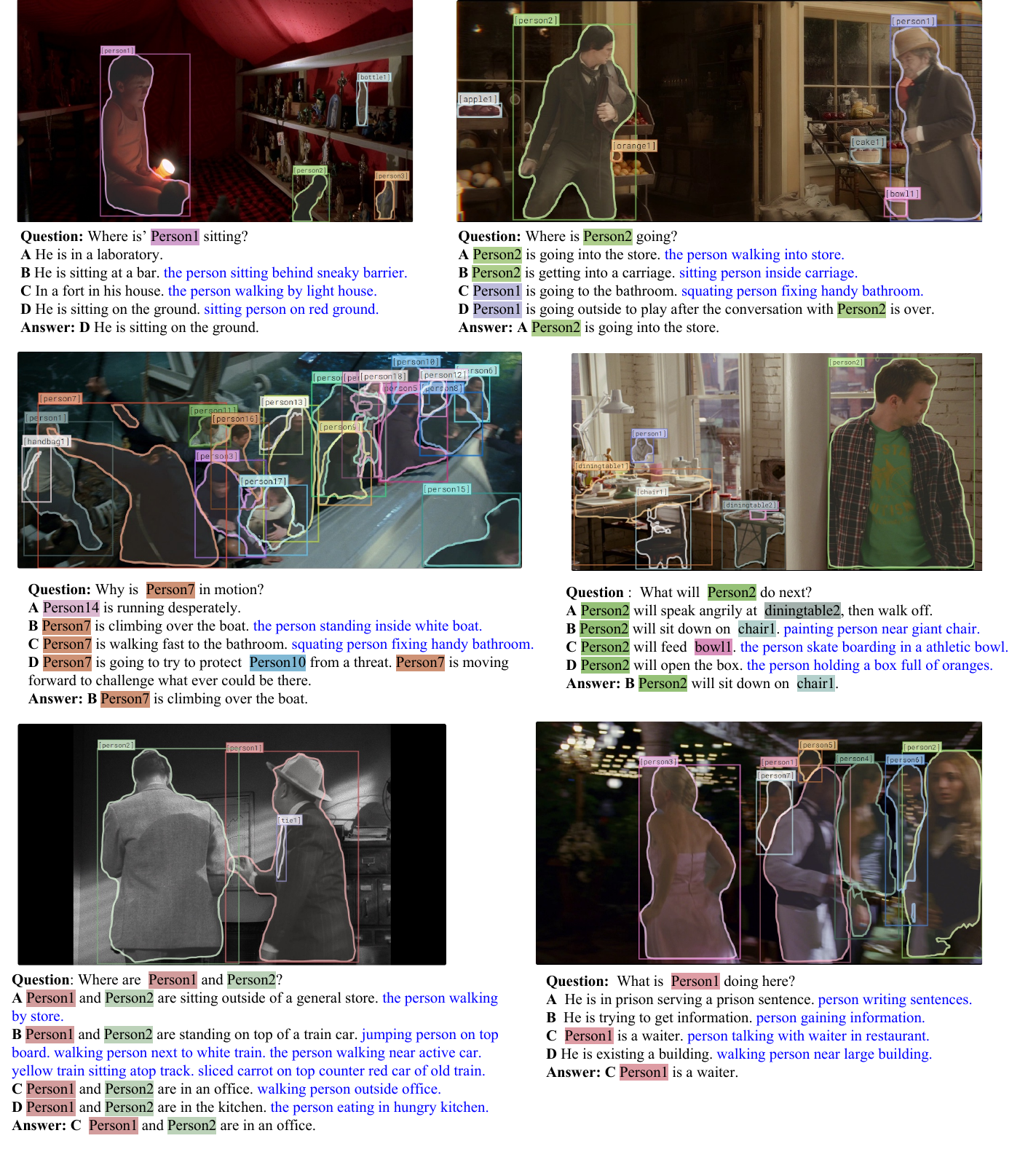}
\caption{Qualitative examples of \ours context enrichment on task VCR.}
\label{fig:vcrapp}
\end{figure}
Figure~\ref{fig:vcrapp} presents more qualitative examples of \ours-based context enrichment in the visual commonsense reasoning (VCR) task.
The context enrichment on VCR is performed at two-level, incorporating both entities and relations: (1) we parse the question and options to obtain all \verb|(S, O)| pairs and, for each entity pair, apply the same relation augmentation as in the image retrieval task; (2) for the \verb|V| in each option, we enrich the visual context using the same method as illustrated in GSR. 
It is shown that unrelated answers are usually enriched with contexts that are not relevant to the image, thus enlarging the distance between incorrect answers and the question, e.g., the enriched contexts ``\textit{squating person fixing handy bathroom}'' for example 3 in Figure~\ref{fig:vcrapp}. At the same time, the knowledge description of the correct answer is enhanced by incorporating information that aligns with the image contents, e.g., the enriched knowledge contexts ``\textit{sitting people on red ground}'' for example 1 in Figure~\ref{fig:vcrapp}.

%%%%%%%%%%%%%%%%%%%%%%%%%%%%%%%%%%%%%%%%%%%%%%%%%%%%%%%%%
\section{Full List of Filtered Verbs for GSR}
\label{app:verb_list}

\begin{table}[h]
\centering
\caption{The full list of filtered verbs for GSR.}
\resizebox{0.95\linewidth}{!}{%
\begin{tabular}{c l}
\toprule
\textbf{Matching Type} & \textbf{The Word List of Event Types} \\
\midrule
\textit{Accurate}
& \multicolumn{1}{m{10cm}}{putting, butting, bathing, dusting, rearing, turning, skating, placing, carting, staring, biting, mashing, folding, wetting, sprinkling, branching, drying, standing, flaming, taxiing, performing, circling, molding, parachuting, glowing, fishing, drinking, speaking, pawing, blocking, milking, racing, stripping, potting, spinning, eating, making, kicking, catching, lacing, urinating, sleeping, pressing, buttering, shearing, sliding, hiking, glaring, dipping, swimming, shopping, slicing, shelling, wagging, grilling, crafting, raining, clawing, splashing, rubbing, snowing, breaking, guarding, clipping, sewing, braiding, telephoning, buttoning, waiting, serving, picking, camping, leaning, working, kissing, wrapping, trimming, tripping, pasting, soaring, driving, kneeling, pumping, coloring, lighting, training, ducking, bowing, arching, cooking, checking, pushing, flipping, rocking, cresting, cleaning, reading, nailing, stitching, building, climbing, covering, shelving, attaching, calming, selling, gluing, dyeing, lapping, photographing, peeling, sprouting, licking, displaying, combing, stacking, planting, fastening, buying, mopping, burning, erasing, measuring, dining, tattooing, gardening, decorating, clearing, fixing, weeding, pulling, feeding, watering, crowning, shaking, dripping, emptying, typing, chasing, poking, leaping, pouring, hanging, sniffing, piloting, falling, overflowing, resting, crashing, carving, ballooning, wading, loading, shaving, boarding, pinning, rowing, juggling, shoveling, hugging, throwing, calling, singing, carrying, walking, writing, crouching, floating, painting, opening, tying, riding, strapping, dialing, saying, bubbling, signing, camouflaging, operating, leading, laughing, parading, skiing, drawing, gnawing, celebrating, spreading, filling, giving, running, smelling, plowing, helping, brushing, scooping, adjusting, wrinkling, steering, biking, smiling, spraying, boating, paying, chewing, stuffing, clinging, landing, wheeling, talking, scoring, teaching, jogging, pitching, flapping, tipping, scrubbing, sitting, surfing, stirring, competing, drumming, jumping, filming, dancing, waxing, hitting, recording, baking, waving, washing, signaling, chopping, stretching, rafting, microwaving, phoning, lifting, swinging, releasing, ramming, towing, packing, hauling, frying 
(\textit{244 words})} \\
\hline
\textit{Fuzzy} & \multicolumn{1}{m{10cm}}{educating, marching, spanking, descending, smearing, heaving, cramming, inflating, stooping, inserting, squeezing, tugging, tilting, moistening, swarming, subduing, waddling, winking, flexing, punching, attacking, nuzzling, sprinting, sucking, puckering, sketching, rotting, videotaping, complaining, tuning, locking, hurling, pricking, arranging, constructing, slapping, sweeping, restraining, dousing, frisking, twisting, wringing, hoisting, immersing, shredding, blossoming, igniting, spying, offering, pouting, confronting, docking, assembling, prying, grinning, sharpening, pruning, disciplining, nipping, coaching, nagging, storming, handcuffing, apprehending, bouncing, clenching, taping, distributing, striking, studying, plunging, curling, aiming, sowing, grinding, rinsing, punting, mowing, hitchhiking, skipping, leaking, providing, hunching, spoiling, kneading, burying, foraging, lathering, vaulting, ejecting, mending, pinching, deflecting, ascending, peeing, bothering, repairing, pedaling, ailing, fueling, skidding, scraping, soaking, grimacing, scolding, spitting, knocking, crushing, bandaging, saluting, fording, stumbling, discussing, raking, launching, whirling, fetching, brawling, retrieving, snuggling, exercising, colliding, stroking, whipping, tilling, betting, farming, browsing, examining, dropping, barbecuing, ignoring, asking, flinging, perspiring, embracing, slipping, flicking, smashing, arresting, lecturing, tearing, gasping, applying, counting, spilling, dragging, recovering, practicing, scratching, shooting, packaging, hunting, stinging 
(\textit{154 words})}\\
\bottomrule
\end{tabular}
}
\label{tab:verb-list}
\end{table}

We provide the full list of verbs out of the predefined 504 candidates of GSR~\cite{pratt2020grounded} that can be accurate-matched or fuzzy-matched to extracted visual knowledge in Table \ref{tab:verb-list}, based on which we compose the testing subset for our evaluation on GSR application in Section 5.2.

\end{document}